\documentclass[10pt,twocolumn,letterpaper]{article}

\usepackage{cvpr}
\usepackage{times}
\usepackage{epsfig}
\usepackage{amssymb}
\usepackage{lipsum}
\usepackage{mathtools}

\usepackage{hyperref}
\usepackage{amsmath,amssymb,amsfonts,mathrsfs}

\usepackage{tikz}
\usepackage{algorithm}
\usepackage{algorithmic}
\usepackage{booktabs}
\usepackage{longtable}
\usepackage{centernot}
\usepackage[capitalise]{cleveref}

\usepackage{caption}
\usepackage{subcaption}

\usepackage{pifont}

\renewcommand{\paragraph}[1]{{\bf #1.}}

\cvprfinalcopy 

\def\cvprPaperID{5} 

\ifcvprfinal\fi

\begin{document}

    \setlength{\abovedisplayskip}{3pt}
    \setlength{\belowdisplayskip}{3pt}

    \title{Hierarchical Image Classification using Entailment Cone Embeddings}
    
    \author{\bf Ankit Dhall$^{1}$, Anastasia Makarova$^{1}$, Octavian Ganea$^{2}$, Dario Pavllo$^{1}$, Michael Greeff$^{1}$, Andreas Krause$^{1}$\\
{$^{1}$ETH Zurich \hspace{1.5cm} $^{2}$MIT}\\
{\tt\small adhall@ethz.ch, anastasiia.makarova@inf.ethz.ch, oct@mit.edu}\\ {\tt\small dario.pavllo@inf.ethz.ch, michael.greeff@usys.ethz.ch, krausea@ethz.ch}
}

	\maketitle
	
    \begin{abstract}
    Image classification has been studied extensively, but there has been limited work in using unconventional, external guidance other than traditional image-label pairs for training. We present a set of methods for leveraging information about the semantic hierarchy embedded in class labels. We first inject label-hierarchy knowledge into an arbitrary CNN-based classifier and empirically show that availability of such external semantic information in conjunction with the visual semantics from images boosts overall performance. Taking a step further in this direction, we model more explicitly the label-label and label-image interactions using order-preserving embeddings governed by both Euclidean and hyperbolic geometries, prevalent in natural language, and tailor them to hierarchical image classification and representation learning. We empirically validate all the models on the hierarchical ETHEC dataset.
\end{abstract}
    \thispagestyle{empty}
    \section{Introduction}
\label{sec:introduction}
\begin{figure}[!htbp]
    \centering
    \includegraphics[width=0.42\textwidth]{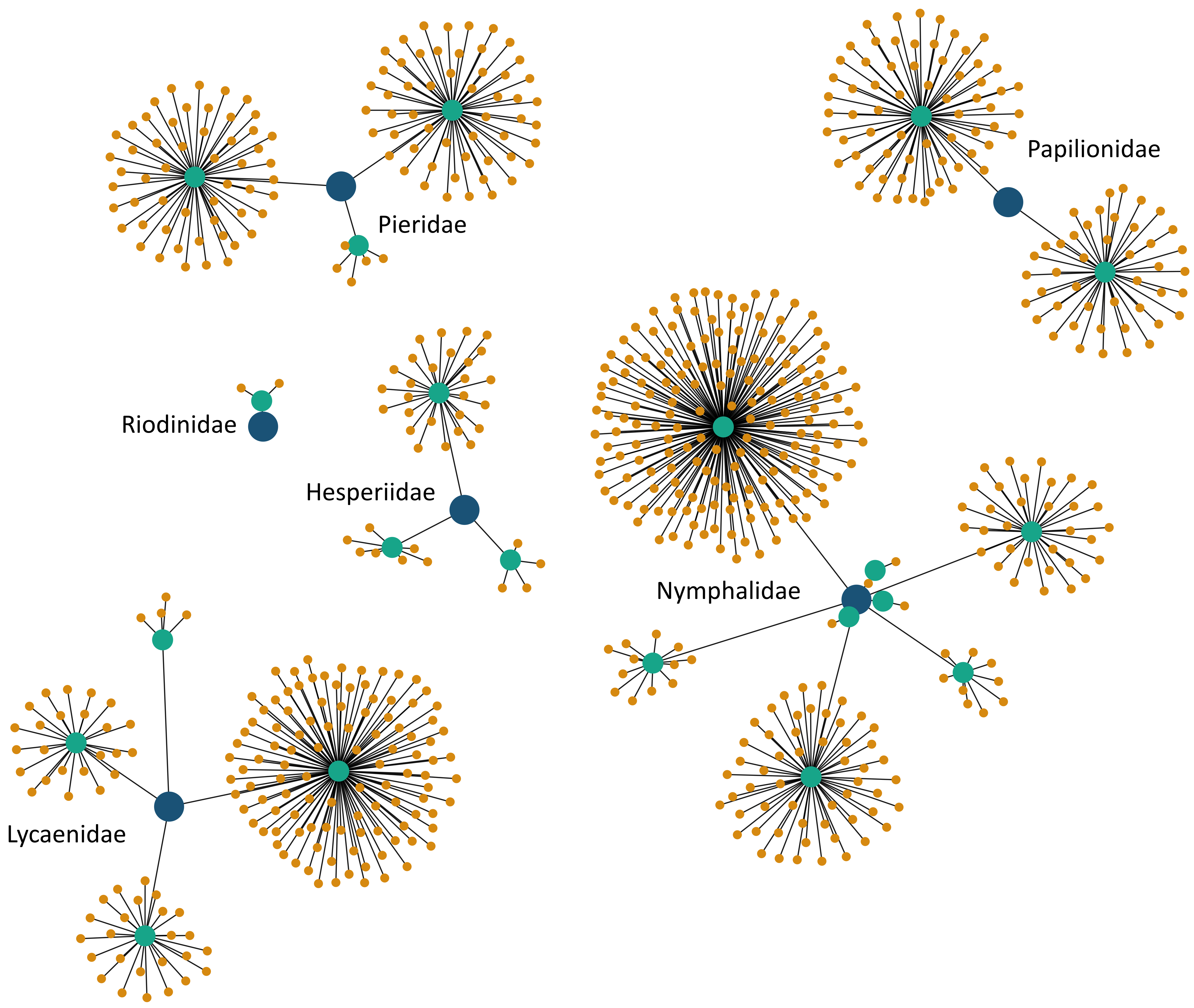}
    \caption{Hierarchy of labels from the ETHEC dataset \cite{dhall_20.500.11850/365379} across 4 levels: family (blue), sub-family (aqua), genus (brown) and species. For clarity, this visualisation depicts only the first 3 levels. The name of the family is displayed next to its sub-tree. Edges represent direct relations.}
    \label{fig:d3_viz}
\end{figure}

In deep learning, classification is typically performed by independently predicting class-probabilities (e.g., using a linear-softmax layer) and predicting the highest scoring label. Such an approach by default assumes mutually exclusive, unstructured labels. Contrary to this assumption, in many common datasets, labels have an underlying latent organization, potentially allowing hierarchical clustering into progressively more abstract concepts. Relatively few previous works use hierarchical information in the context of computer vision. Among them, in \cite{RedmonYOLO9000} the label-hierarchy from WordNet \cite{miller1995wordnet} is used to consolidate data across datasets. \cite{deng2012hedging} show how to optimize the trade-off between accuracy and fine-grained-ness of the predicted label, but their proposed method only considers the semantic similarity and disregards visual similarity. \cite{Samplawski2019} use relation graph information to improve performance over a strong baseline in a zero-shot learning setting. 

Incorporating the hierarchy in the model would improve generalization on classes for which training data is scarce, by leveraging shared features among hierarchically-related classes, e.g.\ ``truck'' and ``car'' both have wheels in their shared superclass ``vehicle''.
As is the case  with \emph{few-shot learning} approaches, sharing information and parameters among the long tail of leaf labels helps overcome this data scarcity problem.


\paragraph{Uncovering the black-box model} If a human is tasked with classifying an image, the natural way to proceed is to identify the membership of the image to abstract labels and then move to more fine-grained labels. Even if an untrained eye cannot tell apart an \emph{Alaskan Malamute} from a \emph{Siberian Husky}, it is more likely to at least get the concept of ``animal'' and its sub-concept ``dog'' correct.

Using the label hierarchy to guide the classification models we are able to bridge one gap in the way machines and humans deal with visual understanding. Incorporating such auxiliary information improves explainability and interpretability of image understanding models.

\paragraph{Leveraging label-label interactions} Usually, image classifiers perform flat N-way classification solely by learning to discriminate between visual signals. These models capture the label-image interactions but do not use additional information available about the inter-label interaction that could boost performance and interpretability.

\paragraph{Long-tailed data distributions} Real-world data is commonly characterized by imbalance. Class labels form a hierarchy and can be viewed as directed acyclic graph (DAG), where abstract labels have finer-grained descendants. Abstract levels have fewer labels and more images per label compared to their fine-grained descendants. The converse is true for fine-grained labels resulting in a long-tailed data distribution. Shallow classifiers benefit from balanced datasets, and generalize worse when classes are imbalanced.  We show that image classifiers can exploit information naturally shared across data from different levels and labels.

\begin{figure}[!htbp]
\centering
\hspace*{-0.5cm}
\includegraphics[width=0.5\textwidth]{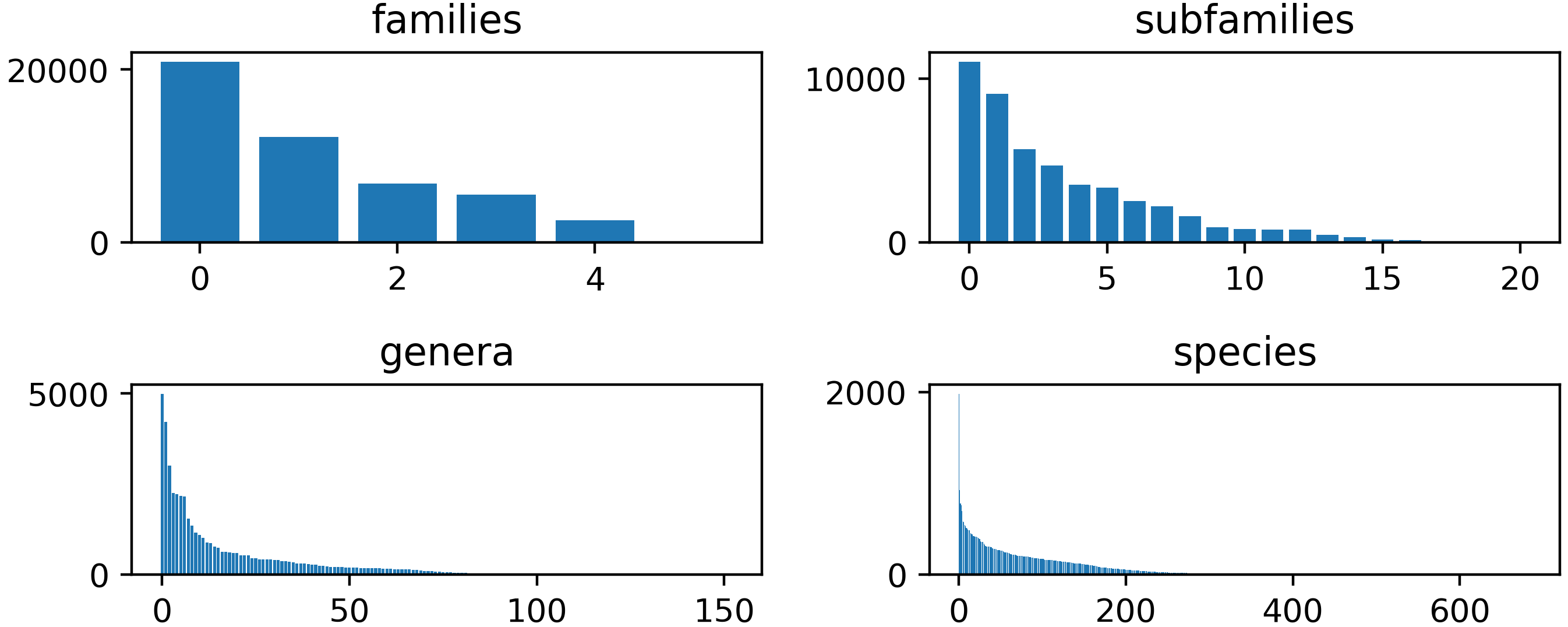}
\caption{Long-tailedness is evident from the image distribution across labels from the 4 levels of our hierarchy: 6 \emph{family}, 21 \emph{sub-family}, 135 \emph{genus} and 550 \emph{species}. x-axis: number of images for a particular label; y-axis: label. Genus and species labels have been omitted for clarity.}
\label{fig:ethec_distribution}
\end{figure}

\paragraph{Visual similarity does not imply semantic similarity} Visual models rely on image-based features to distinguish between different objects. But, often, semantically related classes might exhibit marked visual dissimilarity. Sometimes it might even be the case that the intra-class variance of visual features for a single label is larger than the inter-class variance (we show an example in the Appendix, \cref{fig:semantics-vs-visuals}). In such scenarios learned representations for two instances with different visual appearance would be coerced away from each other, indirectly affecting the image understanding capability of the model.

\begin{figure}[!htbp]
\centering
\includegraphics[width=0.45\textwidth]{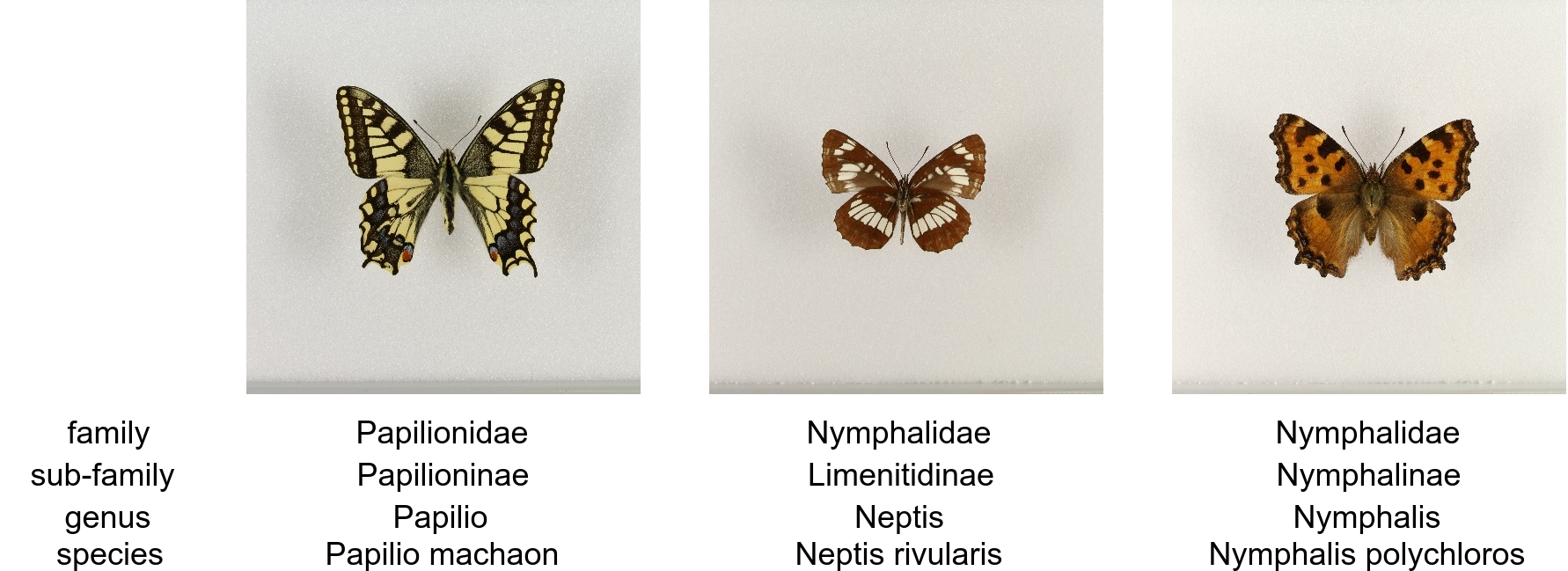}
\caption{Sample images and their 4-level labels from the ETHEC dataset \cite{dhall_20.500.11850/365379}. The dataset consists of 47,978 butterfly specimens with 723 labels spread across 4 levels.}
\label{fig:ethec_samples}
\end{figure}

Labels with varying levels of abstraction may also be beneficial for further downstream tasks involving both natural language and computer vision such as image captioning, scene graph generation and visual-question answering (VQA). This work exploits semantic information available in the form of hierarchical labels. We show that visual models trained with such guidance outperform a hierarchy-agnostic model. We also show how these models can be more interpretable when using more explicit representations via embeddings for the task of image classification.

\paragraph{Our work} We propose and compare multiple approaches for incorporating hierarchical information in state-of-the-art CNN classifiers. To this end, we first compare baselines where the hierarchy is exploited in the loss function (hierarchical softmax, marginalization classifier), and then propose a set of \emph{embedding-based} approaches where images and labels are embedded in a common space. These are more flexible as they allow for entailment prediction tasks and hierarchy-based retrieval. Our embeddings are based on entailment cones, which can be embedded both in Euclidean geometry and in hyperbolic geometry. We compare these and show that the hyperbolic case has empirical advantages over the Euclidean case, while being backed up by theoretical advantages.




We summarize our contributions: (1) applying order-preserving embeddings to image classification, where both images and labels are embedded in a common space that enforces transitivity,  (2) providing a set of methods to incorporate entailment cones in CNN-based classifers, including effective optimization techniques. (3) comparing entailment cones in different geometries (Euclidean and hyperbolic), highlighting their strengths and weaknesses, (4) comparing embedding-based approaches to non-embedding-based approaches, under uniform settings.

	\section{Related Work}
\label{sec:related work}

\begin{figure}
  \hfill
  \centering
  \begin{subfigure}{0.45\columnwidth}
  \includegraphics[width=\textwidth]{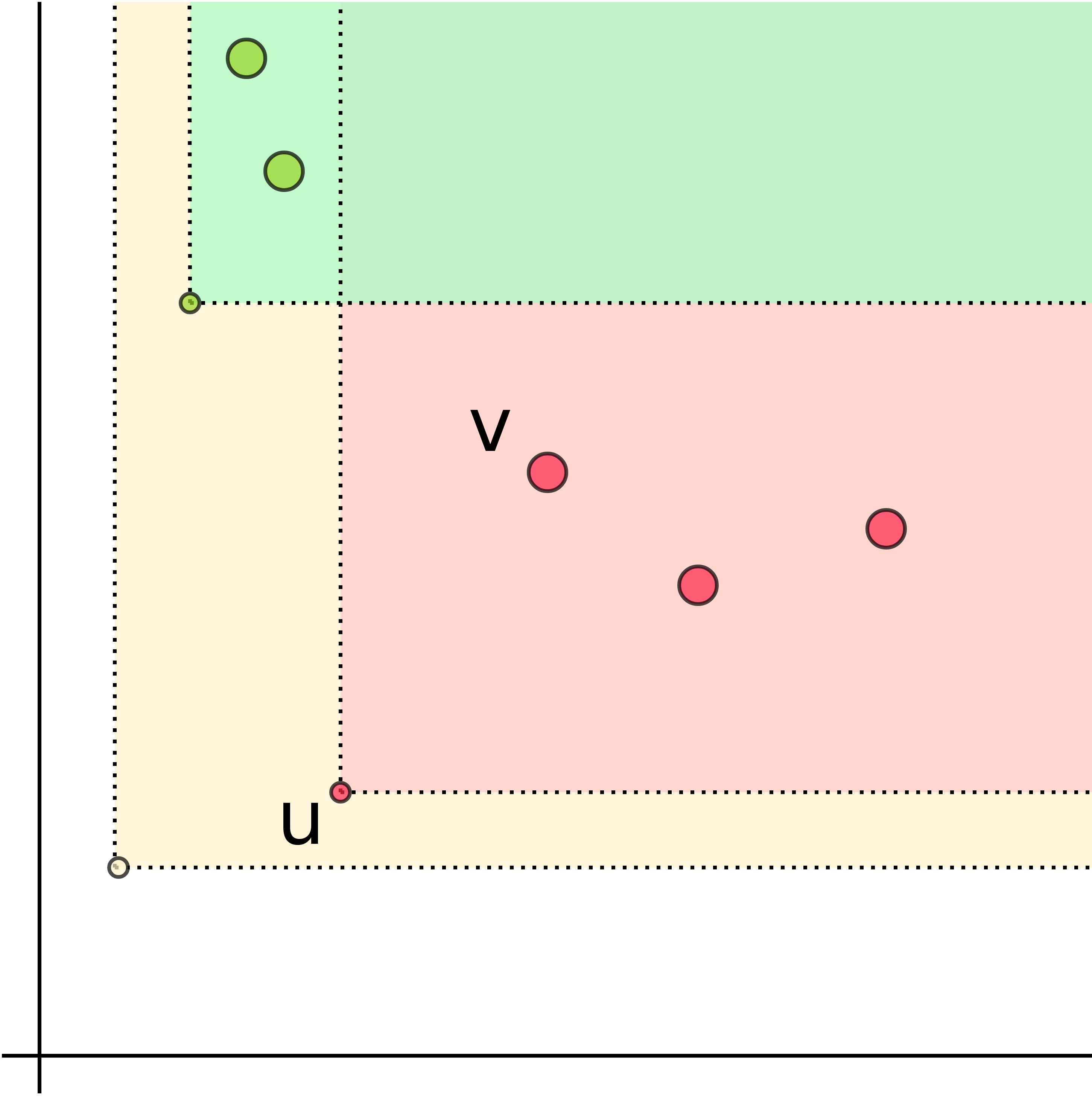}
  \caption{{\small In OE, if $v$ is $u$, it lies within an orthant at $u$.}}
  \end{subfigure}
  \hfill
  \begin{subfigure}{0.45\columnwidth}
  \includegraphics[width=\textwidth]{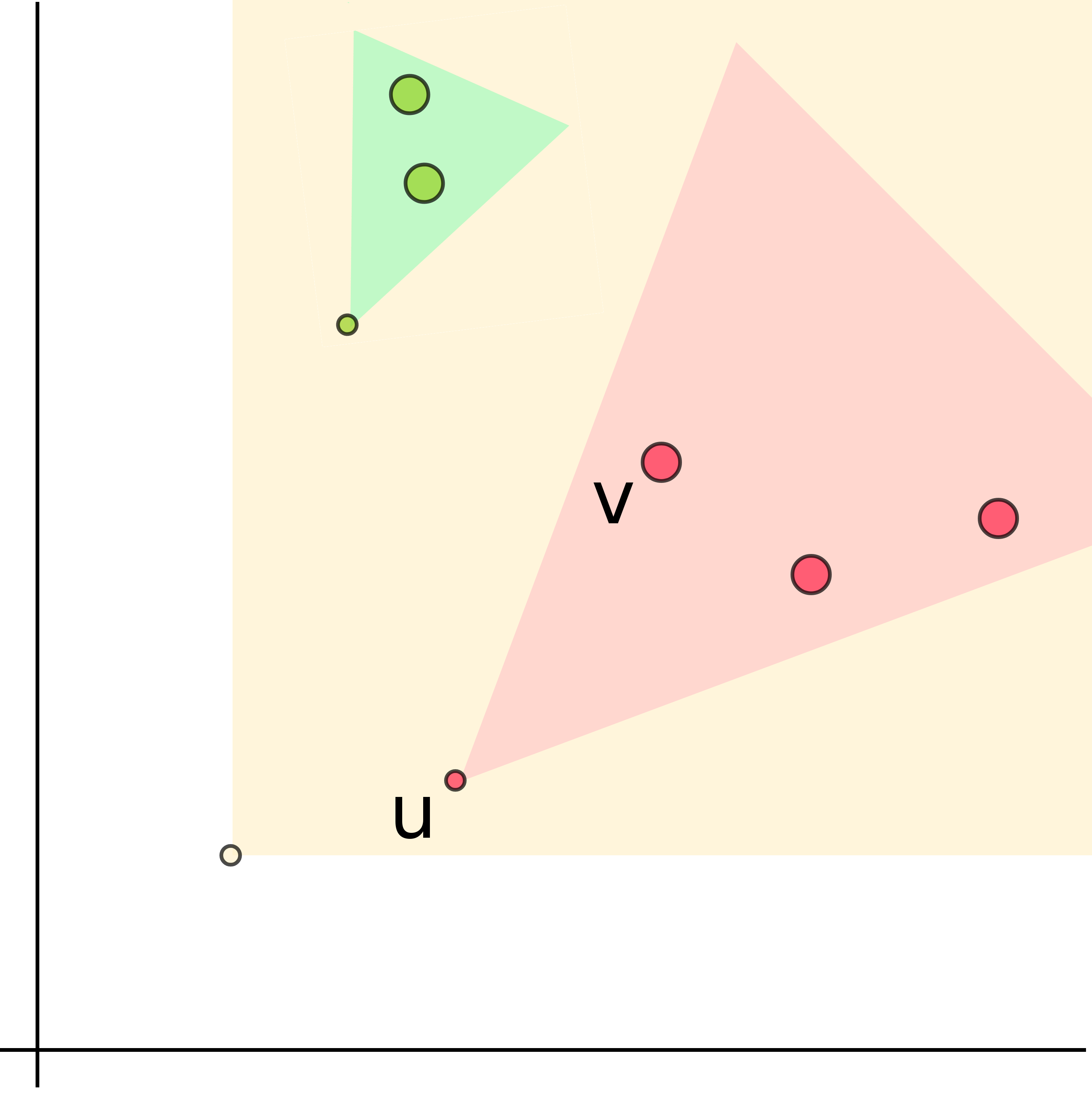}
  \caption{{\small In EC, if $v$ is $u$, it lies within a cone at $u$.}}
  \end{subfigure} 
  \caption{Comparing embedding space for OE and EC.}
  \label{fig:cartoon-representations}
  \hfill
  \vspace{-0.6cm}
\end{figure}

\textbf{Embedding-based models for text.} One way to model semantic hierarchies is to use \emph{order-preserving embeddings}, which enforce transitivity among hierarchically-related concepts by imposing a structure on the latent space. For instance, \emph{order-embeddings} \cite{vendrov2015order} learn hierarchical word embeddings on WordNet \cite{miller1995wordnet}. As an alternative to common symmetric distances (e.g. Euclidean, Manhattan, or cosine), the work proposes an asymmetric distance resulting in the formation of a transitive embedding space as shown in \cref{fig:cartoon-representations}. As opposed to the distance-preserving nature, the order-preserving nature of order-embeddings ensures that anti-symmetric and transitive relations can be captured well without having to rely on physical closeness between points.
However, the distance function in \cite{vendrov2015order} is limited as each concept occupies a large volume in the embedding space irrespective of its volume needs and suffers from heavy orthant intersections. This ill-effect is amplified especially in extremely low dimensions such as $\mathbb{R}^2$. To this end, \cite{ganea2018entailment_cones} proposes \textit{Euclidean entailment cones} which generalizes order-embeddings by substituting  translated orthants with more flexible convex cones. Furthermore, \cite{suzuki2019hyperbolic_disk} generalizes order-embeddings \cite{vendrov2015order} and entailment cones \cite{ganea2018entailment_cones} for embedding DAGs with an exponentially-increasing number of nodes.

More general and flexible methods where the embedding space is not necessarily Euclidean have also been explored. 
\cite{ganea2018entailment_cones} leverage non-Euclidean geometry by learning embeddings defined by \textit{hyperbolic cones} for hypernymy prediction in the WordNet hierarchy \cite{miller1995wordnet}. In hyperbolic space, the volume of a ball grows exponentially with the radius as compared to polynomially in Euclidean space, allowing to embed exponentially-growing hierarchies in low-dimensional space. Lately, \cite{le2019hearst_cones} combined the idea of Hearst patterns to create a graph and hyperbolic embeddings to infer and embed hypernyms from text. \textit{Hyperbolic neural networks} \cite{ganea2018hyperbolicNN} are feed-forward neural networks parameterized in hyperbolic space that allow using hyperbolic embeddings for NLP tasks more naturally and boost the performance.

Other non-Euclidean embeddings include embeddings on surfaces, generalized multidimensional scaling
on the sphere and probability embeddings~\cite{li2018smoothing,muzellec2018} which generalize point embeddings.


\textbf{Embedding-based models for images.}
Visual-semantic embeddings, proposed in \cite{faghri2017vse++}, define a similarity measure instead of an explicit classification and return the closest concept in the embedding space for a given query. They use an LSTM and a CNN and map to a joint embeddings space through a linear mapping and measure similarity for cross-modal image-caption retrieval.  \cite{barz2018hierarchy} maps images onto class embeddings and use dot product to measure similarity. A drawback of such an approach is that the label embeddings are fixed when training on the image embeddings. The labels might be embedded properly however they might not be arranged in a way that puts visually similar labels together. Furthermore, these approaches are based on Euclidean geometry.



In contrast to general CNNs for image classification, the work done in \cite{frome2013devise} exploits unannotated text in addition to the images labels. They use embeddings and transfer knowledge from the text-domain to a model for visual recognition and perform zero-shot classification on an extended ImageNet dataset \cite{deng2009imagenet}.

\textbf{Non-embedding-based approaches.} While this work focuses on embedding-based approaches, there has also been work on incorporating label hierarchies in the model architecture \emph{or} in loss function. 
\cite{kumar2017hierarchical, chen2018finegrained, deng2014large} discuss hierarchical approaches not based on the concept of order-preserving embeddings. While these approaches can effectively exploit label hierarchies to improve performance, their hierarchies are typically fixed, integrated in the architecture of the model, and tailored to one specific downstream task (e.g.\ classification). On the other hand,  embedding-based approaches allow for flexible hierarchies and retrieval tasks using parent-child queries.

\section{Background}

\noindent \textbf{Order-embeddings (OE).} Order-embeddings \cite{vendrov2015order} preserves the \textit{order} between objects rather than distance. From a set of ordered-pairs $\mathcal{P}$ and unordered-pairs $\mathcal{N}$ the goal is to determine if an arbitrary pair is ordered. They use a reversed product order on $\mathbb{R}^{N}$: $y \preceq x \text{ if and only if } \bigwedge_{i=1}^{N} y_{i} \geq x_{i}$ and \textit{approximate} order-violation minimization.

\begin{figure}[!htbp]
    \centering
    \includegraphics[width=0.48\textwidth]{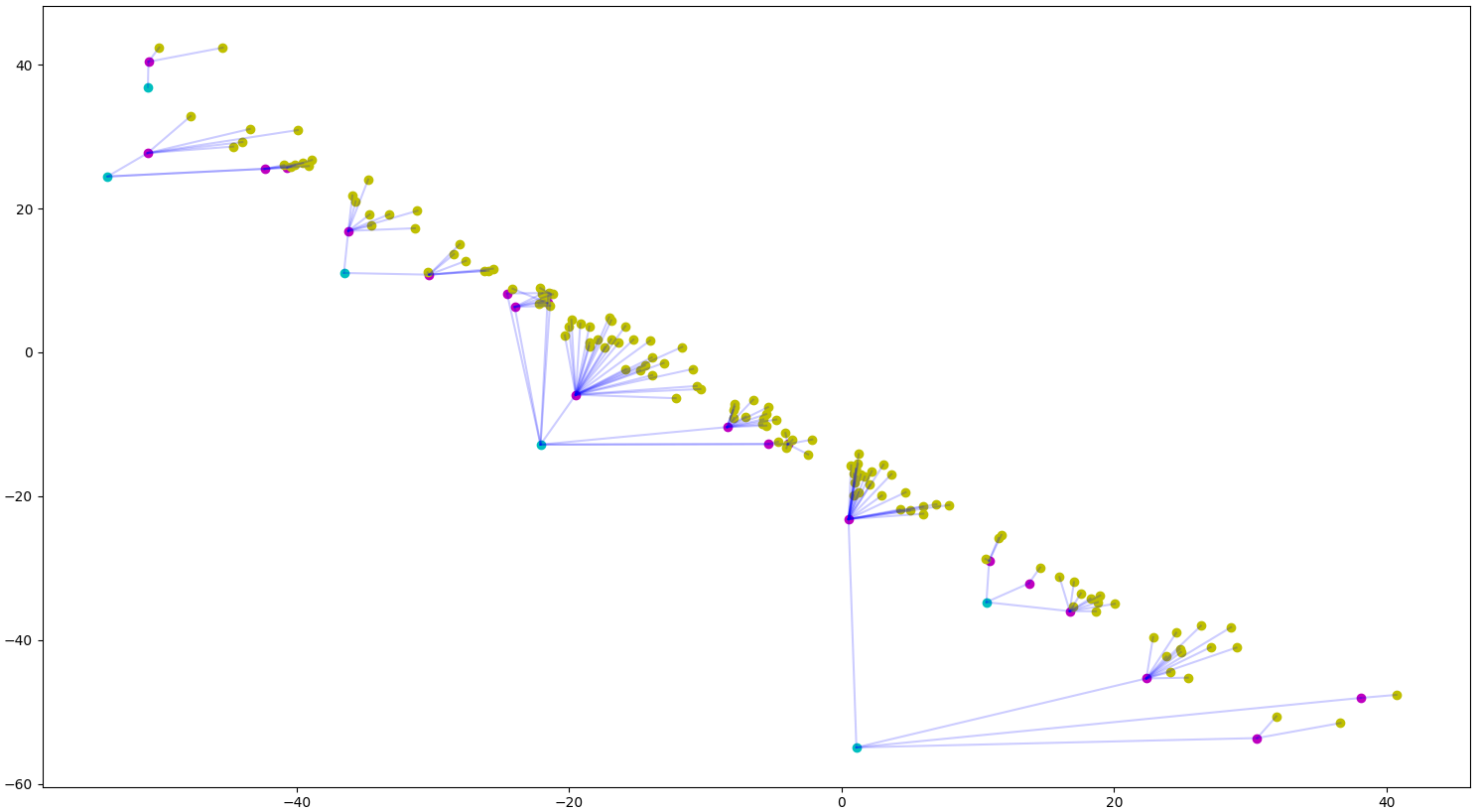}
    \caption{Visualization of the label-hierarchy embedded using OE in $\mathbb{R}^{2}$. Node colors - cyan: \emph{family}, magenta: \emph{sub-family}, yellow: \emph{genus}. Last level omitted for clarity.}
    \label{fig:oe_2d_labels}
\end{figure}

\vspace{-0.5cm}
\begin{equation}
\label{eq:order_embeddings_loss}
\mathscr{L} \mkern-6mu =\mkern-20mu \sum_{(u, v) \in \mathcal{P}}\mkern-13mu  E\big(f(u), f(v)\big)\mkern-5mu + \mkern-7mu \mkern-18mu \sum_{(u', v') \in \mathcal{N}}\mkern-18mu r\big(\alpha \mkern-5mu - \mkern-5mu E(f(u'), f(v'))\big)
\end{equation}

\noindent
where $r(\cdot)\mkern-5mu=\mkern-5mu\max(0,\cdot)$, $\mathcal{P}$ and $\mathcal{N}$ represent positive and negative edges respectively, $\alpha \in \mathbb{R}_{+}$ is a margin, $f$ is a function that maps a concept to its embedding. $E(f(u), f(v))$ is the energy that defines the severity of the order-violation for a given pair $(u, v)$ and is given by $E(x, y) = || \text{max}(0, x-y) ||$. According to the energy $E(x, y) = 0 \iff y \preceq x$. For positive pairs where $y$ \texttt{is-a} $x$, one would like embeddings such that $E(x, y)=0$. $a$ \texttt{is-a} $b$ implies that $a$ is a sub-concept of $b$.

\noindent \textbf{Euclidean Cones (EC).}
Euclidean cones \cite{ganea2018entailment_cones} are a generalization of order-embeddings \cite{vendrov2015order}. For each vector $x$ in $\mathbb{R}^N$, the aperture of the cone is based solely on the Euclidean norm of the vector, $||x||$, \cite{ganea2018entailment_cones} and is given by $\psi(x) = \text{arcsin} (K/||x||)$ where K is a hyper-parameter. The cones can have a maximum aperture of $\pi/2$ \cite{ganea2018entailment_cones}. To ensure continuity and transitivity, the aperture should be a smooth, non-increasing function. To satisfy properties mentioned in \cite{ganea2018entailment_cones}, the domain of the aperture function has to be restricted to $(\varepsilon, 1]$ for some $\varepsilon$. $\varepsilon=f(K)$. \cref{eq:xi} computes the minimum angle between the axis of the cone at $x$ and the vector $y$. $E(x, y) = \text{max}(0, \; \Xi(x, y) - \psi(x))$ measures the cone-violation which is the minimum angle required to rotate the axis of the cone at $x$ to bring $y$ into the cone.

\vspace{-0.4cm}
\begin{equation}
\label{eq:xi}
\Xi(x, y)= \text{arccos}\left(\frac{||y||^2 - ||x||^2 - ||x-y||^2}{2 \; ||x|| \; ||x-y||}\right)
\end{equation}

\begin{figure}[!htbp]
    \centering
    \includegraphics[width=0.47\textwidth]{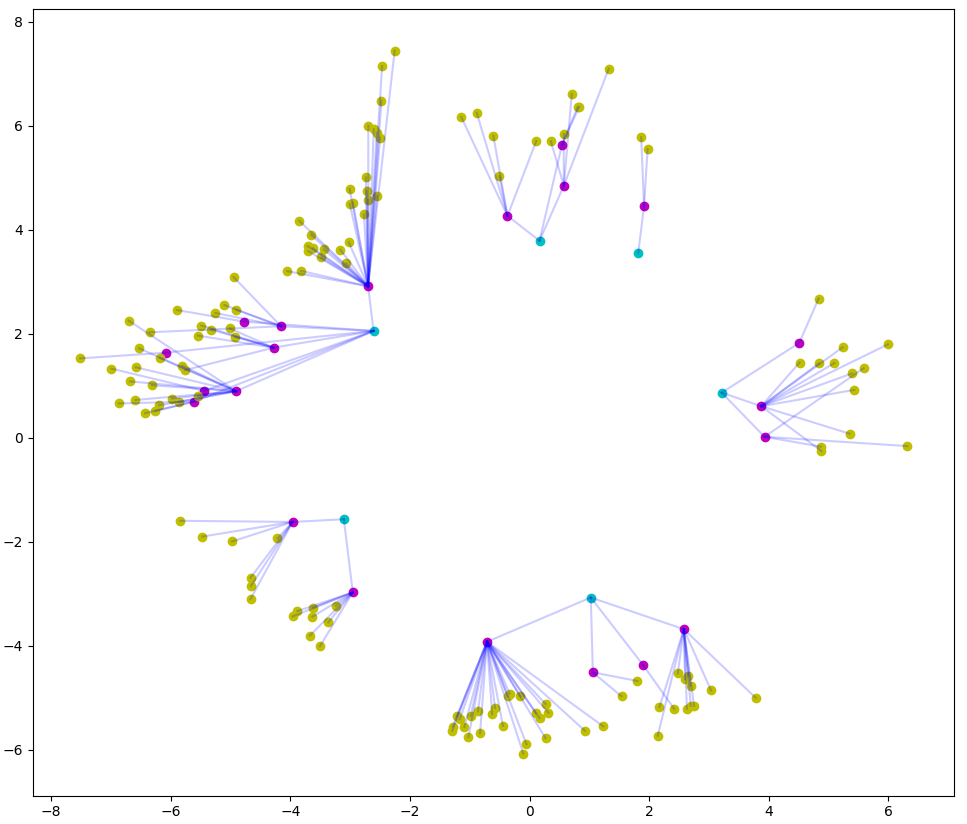}
    \caption{Visualization of the label-hierarchy using Euclidean cones in 2 dimensions. Color coding follows \cref{fig:oe_2d_labels}. \emph{genus+species} nodes are omitted to visualize better.}
    \label{fig:ec_2d_labels}
    \vspace{-0.3cm}
\end{figure}

\noindent \textbf{Hyperbolic Cones (HC).}
\label{subsec:hyperbolic cones math}
The Poincar\'e ball is defined by the manifold $\mathbb{D}^N = \{x \in \mathbb{R}^N: ||x|| < 1\}$. The distance between two points $x, y \in \mathbb{D}^N$ and the norm are $$d_{\mathbb{D}}(x, y) = \text{arccosh} (1 + 2 (||x - y||^2)/((1-||x||^2)(1-||y||^2)))$$ and $||x||_{\mathbb{D}} \mkern-5mu = \mkern-5mu d_{\mathbb{D}}(0, x) \mkern-5mu = \mkern-5mu 2 \, \text{arctanh}(||x||)$ where we use $|| . ||$ for Euclidean norm, $\langle . , . \rangle$ for dot-product and $\hat{x} \mkern-5mu = \mkern-5mu x/||x||$ for a unit vector. The angle between two tangent vectors $u, v \in T_{x}\mathbb{D}^{n}$ is given by $\text{cos}(\angle (u, v)) \mkern-5mu = \mkern-5mu \langle u, v \rangle/(||u|| \; ||v||)$. The aperture of the cone is $\psi(x) \mkern-5mu = \mkern-5mu \text{arcsin} (K (1-||x||^2) / ||x||)$. $\Xi(x, y)$ computes the minimum angle between the axis of the cone at $x$ and the vector $y$.


\vspace{-0.2cm}
\begin{equation}
\label{eq:hyp_cones_xi}
\Xi(x, y) \mkern-5mu = \mkern-5mu \text{arccos} \mkern-5mu \left( \mkern-5mu \frac{\langle x, y \rangle (1 \mkern-5mu + \mkern-5mu ||x||^2) - ||x||^{2}(1 \mkern-5mu + \mkern-5mu ||y||^2)}{\omega \sqrt{1 \mkern-5mu + \mkern-5mu ||x||^2||y||^2 \mkern-5mu - \mkern-5mu 2 \langle x, y \rangle}} \mkern-5mu \right)
\end{equation}

\noindent $E(x, y) \mkern-5mu = \mkern-5mu \max(0, \; \Xi(x, y) - \psi(x))$ measures the cone-violation which is the minimum angle required to rotate the axis of the cone at $x$ to bring $y$ into the cone. $\omega \mkern-5mu = \mkern-5mu ||x|| \; ||x \mkern-5mu -\mkern-5mu y||$

\textbf{Optimization in hyperbolic space.}
\label{subsec:optimization in hyp}
For parameters living in hyperbolic space, Riemannian stochastic gradient descent (RSGD) \cite{ganea2018entailment_cones} is used. An update $u \leftarrow \text{exp}_{u} (\eta \; \nabla_{u}^{R} \mathscr{L})$ involves Rimannian gradient (RG) $\nabla_{u}^{R} \mathscr{L}$ for parameter $u$. RG is computed by rescaling the Euclidean gradient by $\nabla_{u}^{R} \mathscr{L} = (1/\lambda_{u})^2 \nabla_{u} \mathscr{L}$ where $\lambda_{u} = 2/(1 - ||u||^2)$ \cite{ganea2018entailment_cones}.  \textit{Exponential-map} at a point $x$, $\text{exp}_{x}(v): T_{x}\mathbb{D}^{n} \rightarrow \mathbb{D}^{n}$, maps a point $v$ in the tangent space to the hyperbolic space: 

\vspace{-0.4cm}
\begin{equation}
\label{eq:exp_x_v}
\exp_{x}(v) =  (x (c \lambda_{x} + s \langle x, \hat{v} \rangle)) / q + (\hat{v} s)/q
\end{equation}

\noindent where $\lambda_{x}'=(\lambda_{x} - 1)$ and $q=1 + \lambda_{x}'c + \lambda_{x} s \langle x, \hat{v} \rangle$, $s=\text{sinh}(\lambda_{x}||v||)$, $c=\text{cosh}(\lambda_{x}||v||)$.

	\section{Approach}
\subsection{CNN classifiers}
\begin{figure}[!htbp]
    \centering
    \includegraphics[width=0.5\textwidth]{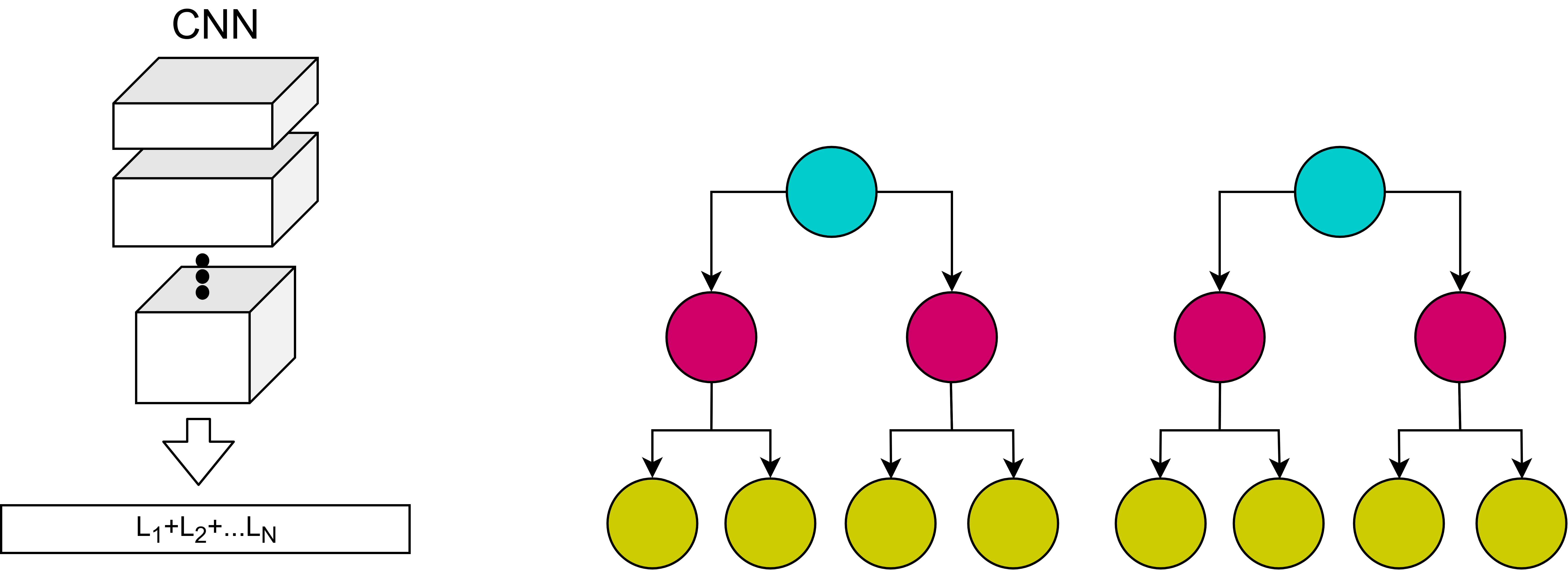}
    \caption{Model schematic for the hierarchy-agnostic classifier. The model is a multi-label classifier and does not utilize any information about the presence of an explicit hierarchy in the labels.}
    \label{fig:multi-label-model-schematic}
\end{figure}

We do not focus on specifically designed CNN components but on different ways to formulate probability distributions to pass hierarchical information. \\
\textbf{Hierarchy-agnostic baseline classifier (HAB).}
As a baseline, we use SOTA residual network for image classification \cite{he2016resnet}. The baseline is agnostic to any label hierarchy in the dataset. The model performs $N_{t}$-way classification (see \cref{fig:multi-label-model-schematic}). $N_{t} = \sum_{i=1}^{L} N_{i}$ represents labels across all $L$ levels and $N_{i}$ are the number of distinct labels on the \textit{i-th} level. It uses the one-versus-rest strategy for each of the $N_{t}$ labels. We minimize multi-label soft-margin loss,

\vspace{-0.4cm}
\begin{equation}
\label{eq:loss_hab}
\mathscr{L}(x, y) = \frac{1}{N_{t}} \sum_{j=1}^{N_{t}} (a_{j} + b_{j})
\end{equation}

\noindent $x \in \mathbb{R}^{N_{t}}$, $y \in \left\{0, 1\right\}^{N_{t}}$. $a_{j}=y_{j} \, \log((1 + \exp(-x_{j}))^{-1})$ and $b_{j}=(1-y_{j}) \, \log(\exp(-x_{j})/(1 + \exp(-x_{j})))$. $\mathcal{F}(\mathcal{I}) = x$, where $x$ are the logits (normalized as a probability distribution) from the last layer of a model $\mathcal{F}$ which takes as input image $\mathcal{I}$. From empirical analysis we found that choosing a single threshold for all labels is better as it is less prone to over-fitting than choosing a per-class decision boundary. Refer to Appendix \ref{subsec:hab_details}.





\begin{figure}[!htbp]
    \centering
    \includegraphics[width=0.5\textwidth]{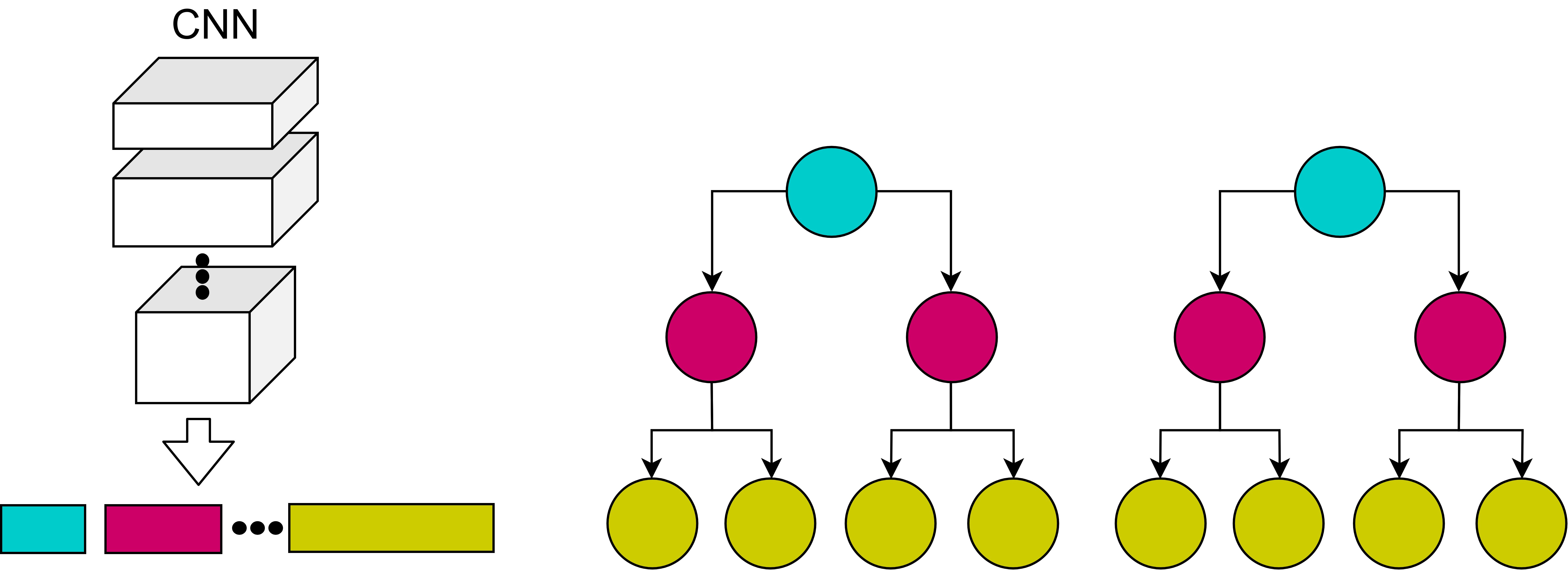}
    \caption{Model schematic for the per-level classifier (=$L$ $N_{i}$-way classifiers). The model use information about the label-hierarchy by explicitly predicting a single label per level for a given image.}
    \label{fig:multi-level-model-schematic}
\end{figure}

\textbf{Per-level classifier (PLC).}
Instead of a single $N_{t}$-way classifier we replace it with $L$ $N_{i}$-way classifiers where each of the $L$ classifiers handles all the $N_{i}$ labels present in level $L_{i}$ (\cref{fig:multi-level-model-schematic}). We use the multi-label soft-margin loss: $\mathscr{L}(x, \tau) = \sum_{i=1}^{L} \mathscr{L}_{i}(x_{i}, \tau_{i})$. 


\vspace{-0.4cm}
\begin{equation}
\label{eq:loss_plc}
\mathscr{L}_{i}(x_{i}, \tau_{i}) = -x_{i}[\tau_{i}] + \log(\sum_{j=1}^{N_{i}} \exp(x_{i}[j]))
\end{equation}

\noindent where, $\tau_{i}$ is the true label for the \textit{i-th} level. $x_{i} \in \mathbb{R}^{N_{i}}$, $\tau \in \mathbb{I}_{+}^L$. $\mathcal{F}(\mathcal{I}) = x$ where, $x$ are the logits from the last layer of $\mathcal{F}$. $x_{i}$ is a continuous sub-sequence of the predicted logits $x$, i.e. $x_{i} = (x_{i}[N_{i-1}+1], x_{i}[N_{i-1}+2], ..., x_{i}[N_{i-1}+N_{i}])$. 

\textbf{Marginalization classifier (MC).}
The notion of $L$ levels is built into the per-level classifier but it is still unaware of the relationship between nodes across levels. Here, a single classifier outputs a probability distribution over the final level in the hierarchy. Instead of having classifiers for the remaining $(L-1)$ levels, we compute the probability distribution over each one of these by summing the probability of the children nodes. Although, the network does not explicitly predict these scores, the models is still penalized for incorrect predictions across the $L$ levels. We minimize $\mathscr{L}(x, \tau) = \sum_{i=1}^{L} \mathscr{L}_{i}(x_{i}, \tau_{i}) = -\sum_{i=1}^{L} \log(p_{i}[\tau_{i}])$ where, $\tau_{i}$ is the true label for the \textit{i-th} level. $x_{i} \in \mathbb{R}^{N_{i}}$, $\tau \in \mathbb{I}_{+}^L$. $\mathcal{F}(\mathcal{I}) = x$ where, $x$ are the logits from the last layer of $\mathcal{F}$.

\vspace{-0.27cm}
\begin{equation}
\label{eq:probability_mc}
p_{i}[j] = P(v_{i}^{j} | \mathcal{I}) = \sum_{c \in \text{childrenOf}(v_{i}^{j})} P(c | \mathcal{I})
\end{equation}
\vspace{-0.33cm}

\begin{figure}[!htbp]
    \centering
    \includegraphics[width=0.5\textwidth]{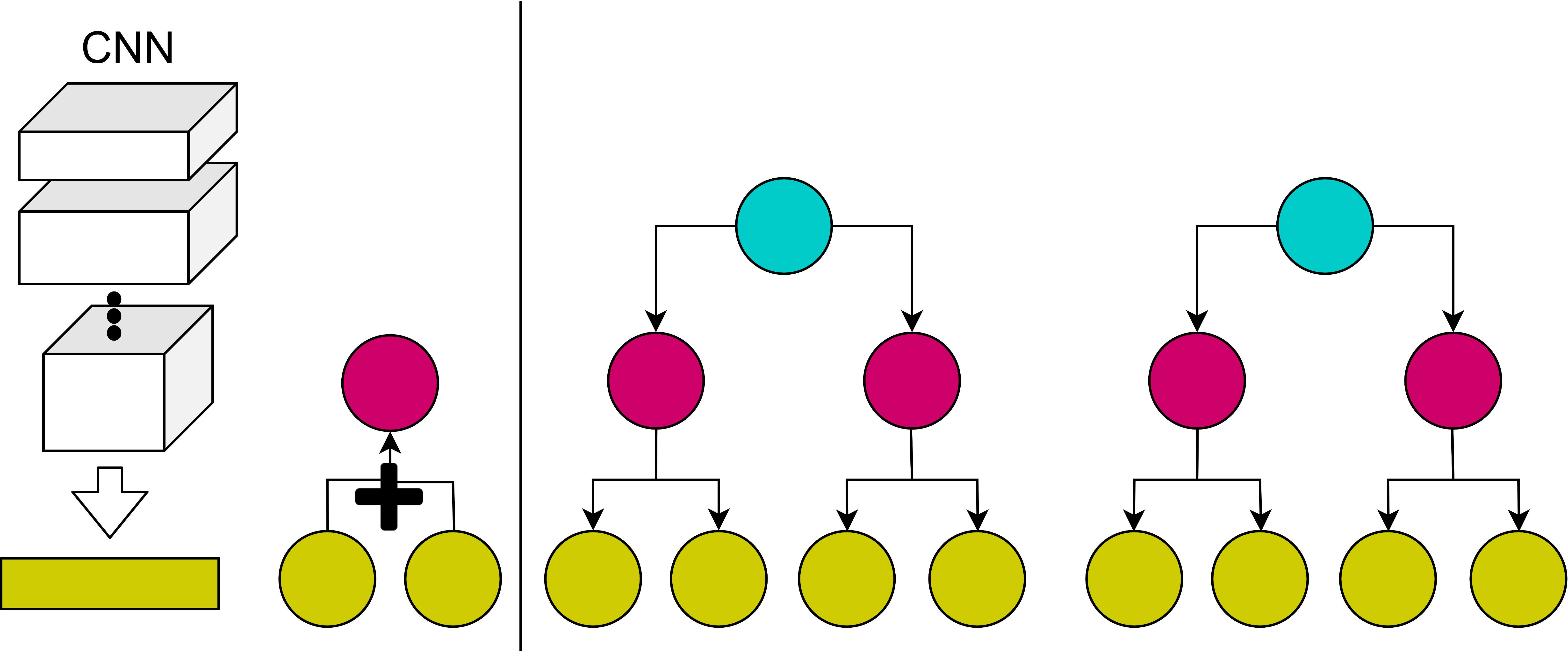}
    \caption{Model schematic for the Marginalization method. Instead of predicting a label per level, the model outputs a probability distribution over the leaves of the hierarchy. Probability for non-leaf nodes is determined by marginalizing over the direct descendants. The Marginalization method models how different nodes are connected among each other in addition to the fact that there are $L$ levels in the label-hierarchy.}
    \label{fig:bs3-marginalization-model-schematic}
\end{figure}

\noindent $\forall i \in \{1, 2, ..., (L-1)\}$ where, $v_{i}^{j}$ is the \textit{j-th} vertex in the \textit{i-th} level. All but the last level use this to compute the probabilities for their labels. For the final level, we compute the probabilities over the leaf nodes by directly using the logits from the model $\mathcal{F}$, using $p_{L}[j] = P(v_{L}^{j} | \mathcal{I}) = \exp(x_{j})/(\sum_{k=1}^{N_{L}} \exp(x_{k}))$. Once $p_{L}$ is determined, $p_{L-1}$ can be calculated in a bottom up fashion as seen in \cref{fig:bs3-marginalization-model-schematic}.


\textbf{Masked Per-level classifier (M-PLC).}
On the upper levels of the hierarchy one has more data per label and fewer labels to choose from. Naturally, this makes classifying relatively accurate closer to the root of the hierarchy. This model exploits knowledge about the parent-child relationship between nodes in a top down manner.

\begin{figure}[!htbp]
    \centering
    \includegraphics[width=0.5\textwidth]{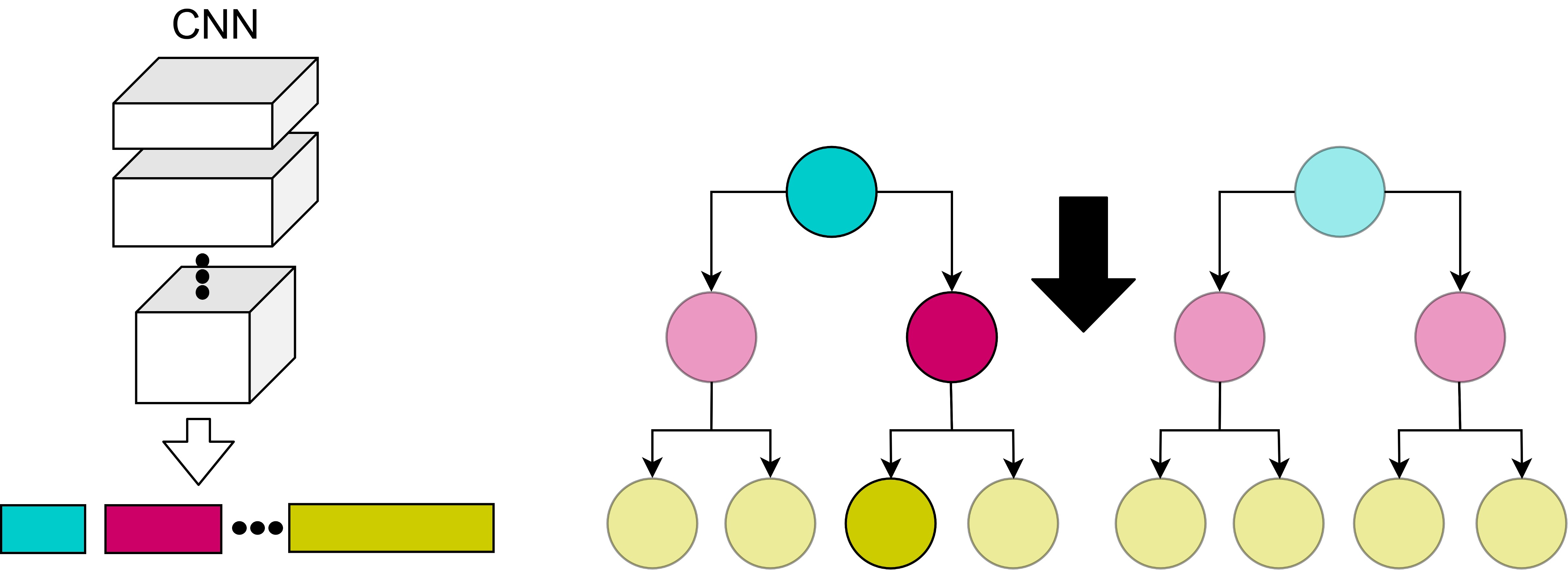}
    \caption{Model schematic for the Masked Per-level classifier. The model is trained exactly like the L $N_{i}$-way classifier. While predicting, one assumes the model performs better for upper levels than lower levels. Keeping this in mind, when predicting a label for a lower level, the model's prediction for the level above is used to mask all infeasible descendant nodes, assuming the model predicts correctly for the level above. This results in competition only among the descendants of the predicted label in the level above.}
    \label{fig:bs3-multi-level-masked-model-schematic}
\end{figure}

Here, we have L-classifiers, one for each level. For level $l_{i}$, the models belief about upper level is leveraged i.e. it's prediction for level $l_{i-1}$. Instead of naively predicting the label with the highest score for level $l_{i}$ (comparing among all possible logits), all nodes except the children of the predicted label for the previous level $l_{i-1}$ are masked (see \cref{fig:bs3-multi-level-masked-model-schematic}). The label for $l_{i}$ is the highest scoring unmasked node. The loss is computed over a subset of the original nodes for any level $l_{i}$ which is possible due to the availability of the parent-child relationship. This assumes that the parent label is correct. Due to less labels and more data, classification in upper levels is more accurate and since we perform this in a top down fashion, this is a reasonable assumption. Another work has shown this to be the case \cite{tasho2018thesis}.

While training, even if the model predicts the parent incorrectly, we still use the ground truth to penalize its prediction for the children. For data with unknown ground truth i.e. during evaluation, the model uses the predictions from level $l_{i-1}$ to infer about level $l_{i}$ by masking nodes that correspond to labels that are not possible as per the hierarchy. We minimize $\mathscr{L}(x, \tau) = \sum_{i=1}^{L} \mathscr{L}_{i}(x_{i}, \tau_{i})$, where

\vspace{-0.3cm}
\begin{equation}
\label{eq:loss_mplc}
\mathscr{L_i}(x_{i}, \tau_{i}) = -x_{i}[\tau_{i}] + \log(\sum_{j \in C} \exp(x_{i}[j]))
\end{equation}
\vspace{-0.3cm}

\noindent $\tau_{i}$ is the true label for the \textit{i-th} level. $x_{i} \in \mathbb{R}^{N_{i}}$, $\tau \in \mathbb{I}_{+}^L$, $C = \text{childrenOf}(v^{\tau_{i-1}}_{i-1})$. $v_{i}^{j}$ is the \textit{j-th} vertex (node) in the \textit{i-th} level and consequently, $v^{\tau_{i-1}}_{i-1}$ is the node corresponding to the ground-truth on level $(i-1)$. $\mathcal{F}(\mathcal{I}) = x$ where, $x$ are the logits from the last layer model $\mathcal{F}$. $x_{i}$ is a continuous sub-sequence of the predicted logits $x$, i.e. $x_{i} = (x_{i}[N_{i-1}+1], x_{i}[N_{i-1}+2], ..., x_{i}[N_{i-1}+N_{i}])$. 

\textbf{Hierarchical Softmax (HS).}
HS model predicts logits for every node in the hierarchy. There are dedicated linear layers for each group of sibling nodes leading to a separate (conditional) probability distribution over them. This is probability conditioned on the parent node i.e. $p(v_{i}^{j_{i}}|v_{i-1}^{j_{i-1}}), \forall v_{i}^{j_{i}} \in C$, such that $C = \text{childrenOf}(v_{i-1}^{j_{i-1}})$.

To reduce computation over large vocabularies, \cite{morin2005hierarchical, mikolov2013hierarchical2} propose similar ideas for NLP. In the context of computer vision it is relatively unexplored and we propose to predict conditional distributions for each set of direct descendants to exploit the label-hierarchy. \\

\vspace{-0.8cm}
\begin{equation}
\label{eq:probability_hs}
p(v_{i}^{j_{i}}|v_{i-1}^{j_{i-1}}) \mkern-4mu = \mkern-4mu \exp(x_{v_{i-1}^{j_{i-1}}}[j_{i}]) / (\sum_{k \in C} \exp(x_{v_{i-1}^{j_{i-1}}}[k]))
\end{equation}
\vspace{-0.3cm}

\noindent $\forall v_{i}^{j_{i}} \in C$, $x_{v_{i-1}^{j_{i-1}}} \in \mathbb{R}^{|C|}$. The vector $x_{v_{i-1}^{j_{i-1}}}$ represents the logits that exclusively correspond to all the children of node $v_{i-1}^{j_{i-1}}$. With this in place, for each set of children of a given node, a conditional probability distribution is output by $\mathcal{F}$. $\mathcal{F}(\mathcal{I}) = p(\cdot)$ where, $p(\cdot)$ is the conditional probability for every child node given the parent, $p(v_{i}^{j_{i}}|v_{i-1}^{j_{i-1}})$. In order to calculate the joint distribution over the leaves, probabilities along the path from the root to each leaf are multiplied as $p(v_{1}^{j_1}, v_{2}^{j_2}, ..., v_{(L-1)}^{j_(L-1)}, v_{L}^{j_L}) = p(v_{1}^{j_1})p(v_{2}^{j_2}|v_{1}^{j_1})... p(v_{L}^{j_L}|v_{(L-1)}^{j_(L-1)})$ where, $v_{i}^{j_i}$ is the parent node of $v_{i+1}^{j_{(i+1)}}$. The nodes belonging to the i\textit{-th} level and the (i+1)\textit{-st} level respectively.



The cross-entropy loss is computed only over the leaves but since the distribution is calculated using internal nodes, all levels are optimized implicitly. $\mathscr{L}(x, \tau) = -\log(p(v_{1}^{j_1}, v_{2}^{j_2}, ..., v_{(L-1)}^{j_(L-1)}, v_{L}^{\tau_{L}})) = -\log(p(v_{1}^{\tau_{1}}, v_{2}^{\tau_{2}}, ..., v_{(L-1)}^{\tau_{L-1}}, v_{L}^{\tau_{L}}))$, where, $\tau_{i}$ is the true label for the \textit{i-th} level. $x_{i} \in \mathbb{R}^{N_{i}}$, $\tau \in \mathbb{I}_{+}^L$.


\subsection{Embedding Classifiers}
We treat our label hierarchy as a directed-acyclic graph, more specifically as a directed tree graph. The dataset $\mathcal{X}$ consists of entailment relations $(u, v)$ connected via a directed edge from $u$ to $v$. (following the definition in \cite{ganea2018entailment_cones}). These directed edges or hypernym links convey that $v$ is a sub-concept of $u$.

\subsubsection{Label and Image Representations}
\textbf{Label embeddings.} For our implementation of the HC, the label-embeddings live in the hyperbolic space $\mathbb{D}^N$ and are optimized using the RSGD as per \cref{subsec:optimization in hyp}. RSGD is implemented by modifying the SGD gradients in PyTorch\cite{pytorch} as it is not a part of the standard library.

\noindent \textbf{Image embeddings.} For images, features from the final layer of the backbone of the best performing CNN-based model are used ($\in \mathbb{R}^{2048}$). In order to map them to $\mathbb{D}^N$ we use a linear transform $W \in \mathbb{R}^{2048 \times N}$ and then apply a projection into $\mathbb{D}^N$ via the exponential-map at zero which is equivalent to $\text{exp}_{0}(x)$. This bring the image embeddings to the hyperbolic space with Euclidean parameters. This allows for optimizing the parameters with well know optimization schemes such as Adam \cite{kingma2014adam}.

\subsubsection{Embedding Label-Hierarchy}
\label{subsec:embedding_label_hierarchy_only}
We begin by learning to represent the taxonomical hierarchy alone. Considering only the label-hierarchy and momentarily excluding the images we model this problem as hypernym prediction where a hypernym pair represents two labels $(x, y)$ such that $y$ \texttt{is-a} $x$. Embeddedings for the label-hierarchy with OE and EC are shown in \cref{fig:oe_2d_labels} and \cref{fig:ec_2d_labels}.

\textbf{Data splitting.} We use the tree to form the ``basic'' edges for which the transitive closure can be fully recovered. If these edges are not present in the \emph{train} set, the information about them is unrecoverable and therefore they are always included in the \emph{train} set. Now, we randomly pick edges from the transitive closure \cite{wiki:transitive_closure} minus the ``basic'' edges to form a set of ``non-basic'' edges. We use the ``non-basic'' edges to create \emph{val} (5\%) and \emph{test} (5\%) splits and a proportion of the rest are reserved for training. 

\textbf{Training details.} We follow the training details in \cite{ganea2018entailment_cones}. We augment both the validation and test set by 5 negative pairs each for $(x, y)$: of the type $(x', y)$ and $(x, y')$ with a randomly chosen edge that is not present in the full transitive closure of the graph. Generating 10 negatives for each positive. We report performance on different training set sizes. We vary the training set to include 0\%, 10\%, 25\%, 50\% of the ``non-basic'' edges selected randomly. We train for 500 epochs with a batch size of 10. We run two sets of experiments: one, we fix $\alpha=1.0$ as mentioned in \cite{vendrov2015order} and two, tune $\alpha$ based on the F1-score on the \emph{val} set \cite{ganea2018entailment_cones}.

\textbf{\emph{Pick-per-level} strategy.} During the experiments, instead of sampling a negative edge $(x', y)$ uniformly from candidate $x'$, we pick each $x'$ from a different level in the hierarchy. This serves a dual purpose. 78.24\% of the nodes belong to the final level in the hierarchy and uniform negative sampling would result in edges where $x'$ is from the last level majority of the times, making convergence slow. Secondly, this strategy samples hard negatives edges from the same level as the non-corrupted node $y$, helping embeddings to disentangle and spread out in space.

\begin{figure*}[!htbp]
    \centering
    \begin{subfigure}[b]{0.480\textwidth}
        \centering
        \includegraphics[width=\textwidth]{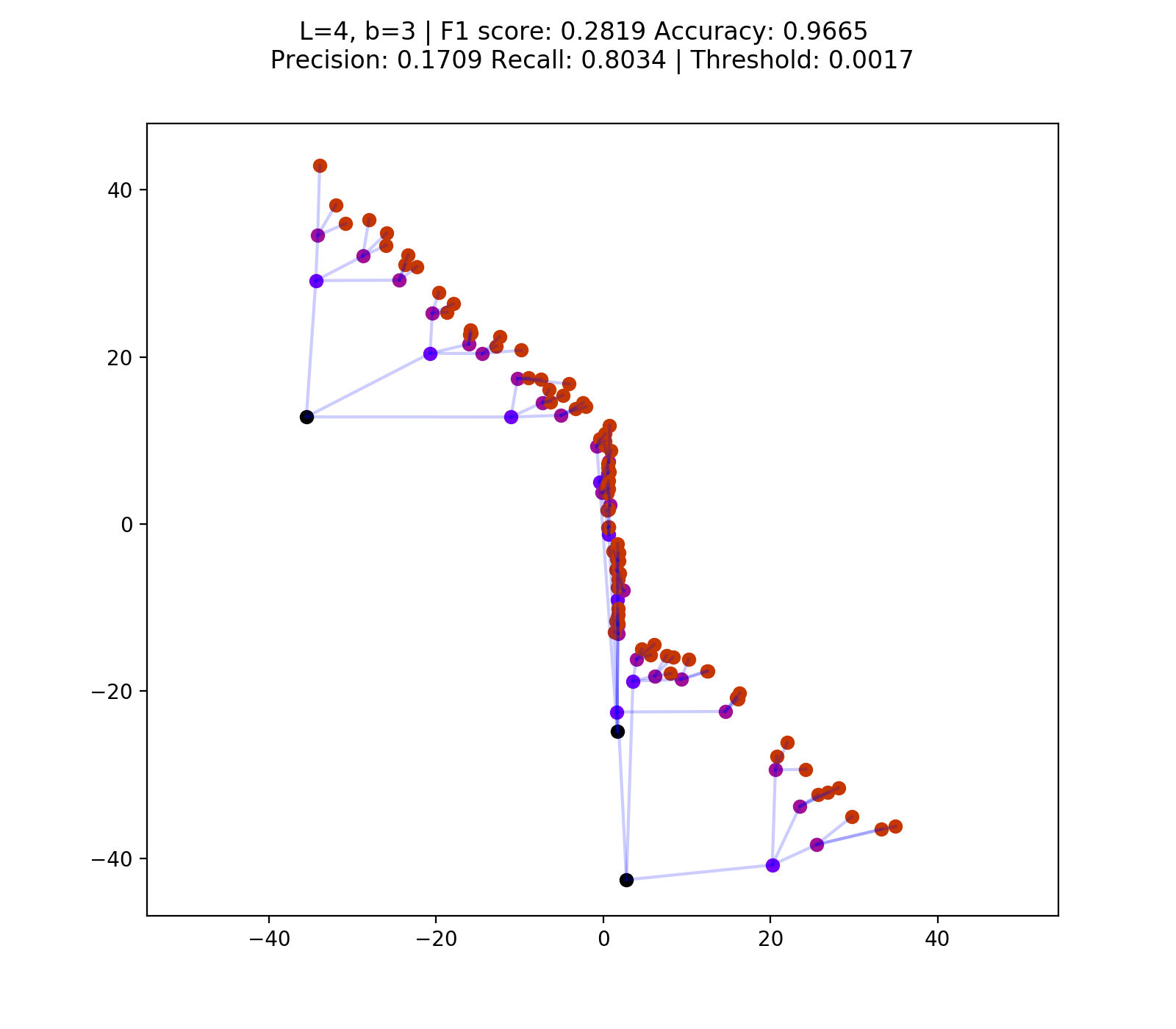}
        \caption[]%
        {{\small Order-embeddings L=4, b=3}}    
        \label{fig:oe_l4b3}
    \end{subfigure}
    \hfill
    \begin{subfigure}[b]{0.480\textwidth}  
        \centering 
        \includegraphics[width=\textwidth]{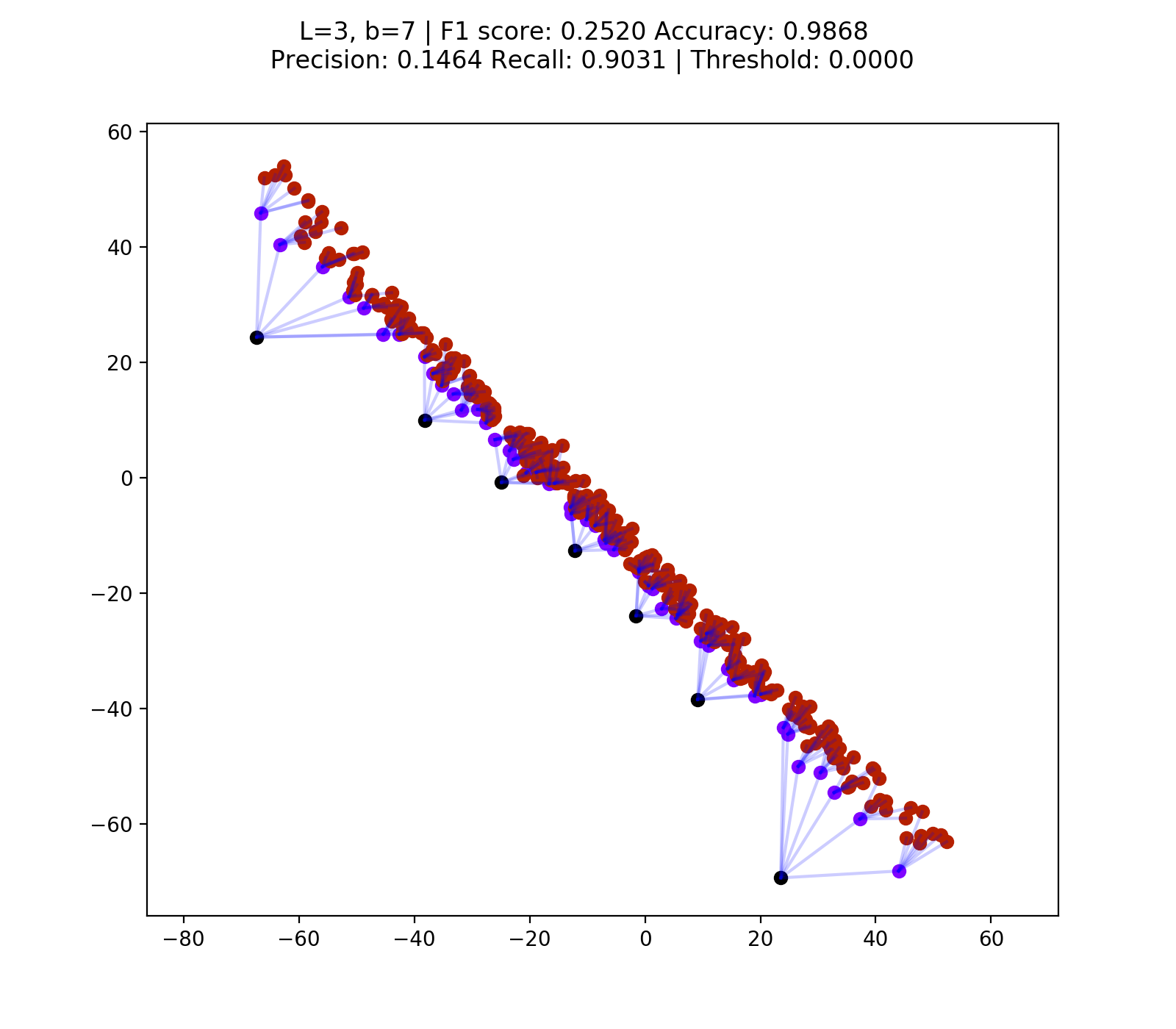}
        \caption[]%
        {{\small Order-embeddings L=3, b=7}}    
        \label{fig:oe_l3b7}
    \end{subfigure}
    \vskip\baselineskip
    \begin{subfigure}[b]{0.480\textwidth}   
        \centering 
        \includegraphics[width=\textwidth]{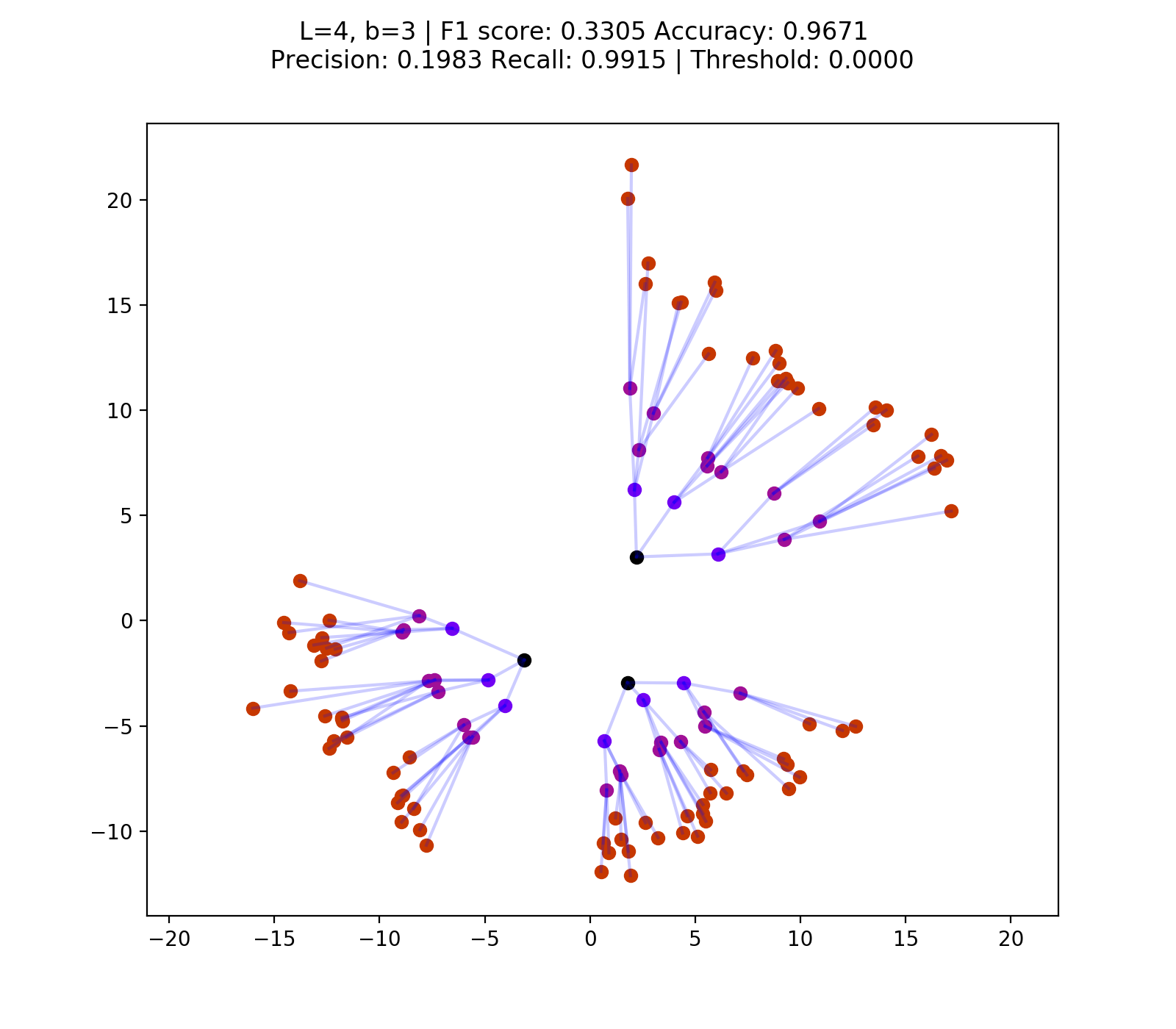}
        \caption[]%
        {{\small Euclidean cones L=4, b=3}}    
        \label{fig:ec_l4b3}
    \end{subfigure}
    \quad
    \begin{subfigure}[b]{0.480\textwidth}   
        \centering 
        \includegraphics[width=\textwidth]{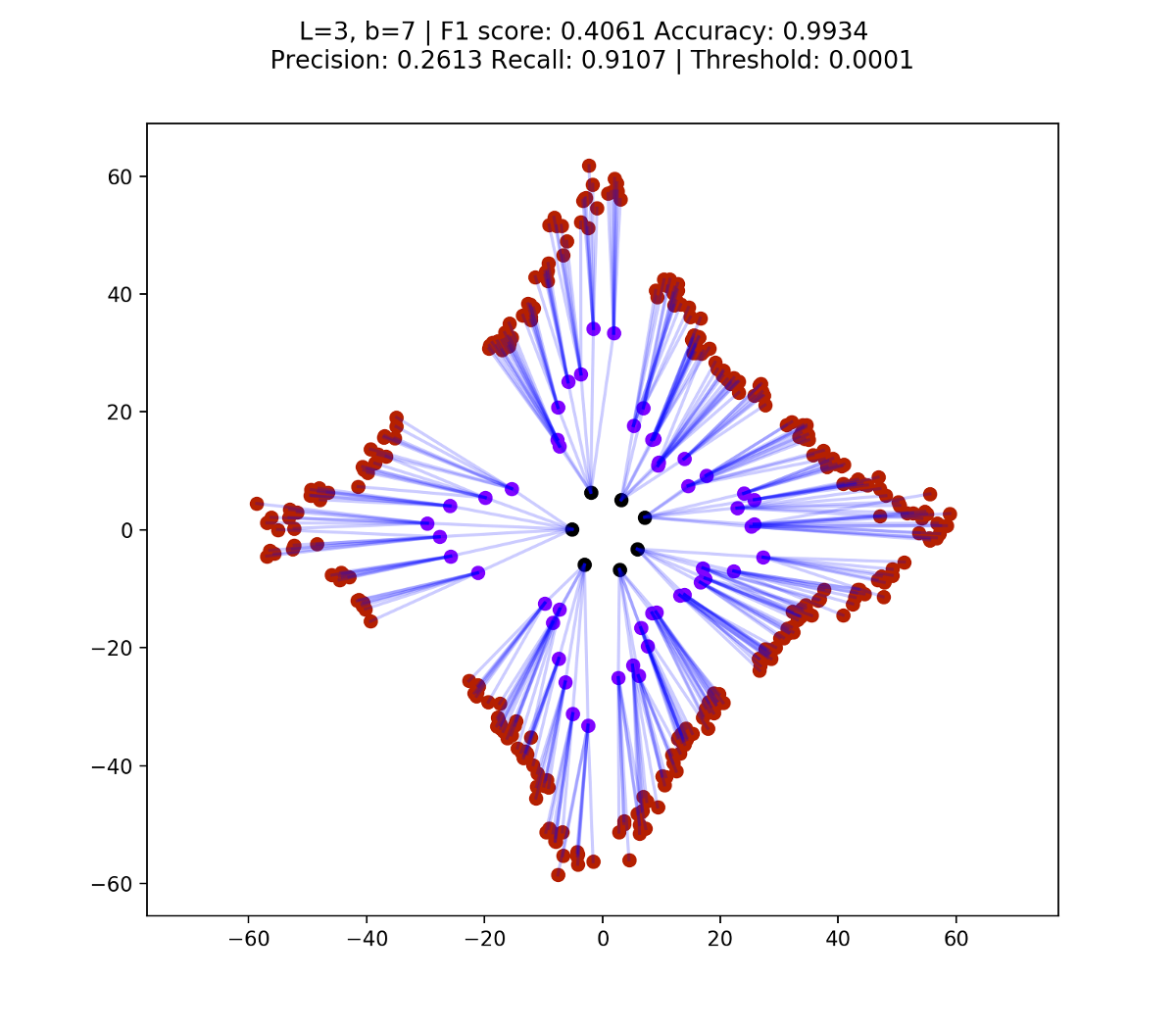}
        \caption[]%
        {{\small Euclidean cones L=3, b=7}}
        \label{fig:ec_l3b7}
    \end{subfigure}
    \caption[]
    {We embed 2 different toy graphs. One with 4 levels and a branching factor of 4 and another one with 3 levels and a branching factor of 7. The model is trained for 1000 epochs with Adam (learning rate of 0.01). The toy graphs are embedded using both order-embeddings and euclidean cones in $\mathcal{R}^2$. We draw an edge between each node that is connected in the original in order to better visualize the embedding quality. Nodes from different levels are colored differently. The illustrations show the levels and branching factor, the edges are split into \emph{train}, \emph{val} and \emph{test} and report F1-score, precision, recall and accuracy; and the threshold to decide if a pair of nodes have a directed edge or equivalently if they are hypernyms.} 
    \label{fig:toy_trees}
\end{figure*}

\textbf{Optimization details.}
We use Adam optimizer \cite{kingma2014adam} for order-embeddings and Euclidean cones. For hyperbolic cones we use RSGD \cite{ganea2018entailment_cones}. $lr=0.01$.
We also embed synthetic trees of varying height and branching factor using OE and EC. The final embeddings are visualized in \cref{fig:toy_trees}.
\subsubsection{Jointly Embedding Images with Label-Hierarchy}
In order-embeddings \cite{vendrov2015order}, the images are put on the lower-level and the captions on the upper level as images are more detailed while captions represent concepts more abstract than the image itself. 
For jointly embedding the images together with the labels we use the hypernym loss from \cref{eq:order_embeddings_loss}. We modify it such that now in addition to the labels,  $\mathcal{G}$ (the graph representing the hierarchy) also contains images as nodes as leaves at the lowest level. $\mathcal{G}$ constitutes of two types of edges: an edge $(u, v)$ can be such that $u, v \in \text{labels}$  or $u \in \text{labels}, v \in \text{images}$. The embeddings are computed differnently for images and labels but in the end, both $f_i$ and $f_l$ map respective inputs to the same space. \\
\textbf{Multi-label Classification with Embeddings}
Since our problem does not concern hypernym prediction but rather assigning multiple labels to an image; instead of performing edge prediction (as the case would be in a hypernym prediction task) we use the embeddings for the task of classification.
To classify an image we compute the order-violation energy $E$ between the given image and each label and pick the label corresponding to the minimum violation, $\text{arg}\,\min\limits_{l}\, E(f_{l}(l), f_{i}(i)), \forall l \in \text{labels}$. \\
\textbf{Generating Label and Image Embeddings} To generate image embeddings we use the best performing CNN model trained on the ETHEC dataset and extract \textit{fc7}-features from the penultimate layer. We use a learnable linear transformation, a matrix $W$, on top of the \textit{fc7}-features to be able to adjust the \textit{fc7}-features and map them into the joint embedding space: $f_{i}(i) = W * \text{CNN}(i) \in \mathbb{R}^{N}$.
$\text{CNN}(i)$ represent the \textit{fc7}-features from our best performing CNN model and $W$ is a matrix. The weights of the CNN are frozen to calculate the \textit{fc7}-features with only $W$ that can be learned. For the labels, $f_{l}(l)$ is just a lookup table that stores vectors in $\mathbb{R}^{N}$. The embedding are in $\mathbb{R}^{N}$ for Euclidean models and $\mathbb{D}^{N}$ for hyperbolic models (Poincar\'e disk). 

\textbf{Data splitting.} We split the data the same way as for the CNN models: \emph{train} (80\%), \emph{val} (10\%) and \emph{test} (10\%) based solely on the images. The graph $\mathcal{G}$ contains directed edges from each label to the image that it ``describes'' as well as edges between related labels.\\
\textbf{Training details.} Let $\mathcal{G}$ represent the graph to be embedded. All edges in $\mathcal{G}_{tc}$, the transitive closure of $\mathcal{G}$, are considered as positive edges. To obtain negative edges, $\mathcal{G}_{neg}$ is constructed by removing the edges in $\mathcal{G}_{tc}$ from a fully-connected di-graph with the same nodes as $\mathcal{G}$.

While training, we generate negative pairs as mentioned in \cref{subsec:embedding_label_hierarchy_only} with the \emph{pick-per-level} strategy. We make sure that we do not sample a negative edge $(u',v')$ such that both $u$ and $v$ are images. This ensures that no two images are forced apart unless their labels require them to do so. For validation and testing, we measure the model's classification the \emph{val} and \emph{test} set images respectively.


\textbf{Graph reconstruction task.} In addition to the classification task, we also check the quality of reconstruction of the label-hierarchy itself. Here, all the edges in $\mathcal{G}$ that correspond to edges between labels are treated as positive edges, while the the edges in $\mathcal{G}_{neg}$ that correspond to edges between labels are treated as negative edges. We compute $E(u, v)\; \forall e \in \mathcal{P} \cup \mathcal{N}$ where $e=(u, v)$ and choose a threshold to classify edges as positive and negative using that yields the best F1-score on this label-hierarchy reconstruction task. This task does not use any edges that have an image on any side to check the quality of reconstruction.

For $W$ we use a linear transformation, a matrix $\mathbb{R}^{2048 \times N}$. Non-linearity is not applied to the output that maps to the embedding space.

\textbf{Optimization details.}
For jointly embedding labels and images, we empirically found using Adam \cite{kingma2014adam} optimizer instead of the RSGD. The label embeddings are parameterized in the Euclidean space and we use the $\text{exp}_0(v)$ to map them to the hyperbolic space. This is observed to be more stable and helps better converge the joint embeddings. Also, with this implementation of the hyperbolic cones, for both labels and joint embeddings, it was not necessary to initialize the embeddings with the Poincar\'e embeddings \cite{nickel2017poincare} as suggested in \cite{ganea2018entailment_cones}. However, a performance boost is obtained when initialized with values from embedding only the label-hierarchy. EC: 200 epochs, $lr_{labels} \mkern-5mu = \mkern-5mu 10^{-2}$, $lr_{im} \mkern-5mu = \mkern-5mu 10^{-3}$. HC: 100 epochs, $lr_{labels} \mkern-5mu = \mkern-5mu 10^{-4}$, $lr_{im} \mkern-5mu = \mkern-5mu 10^{-3}$, Initialization from label-embeddings only model. Adam and $\alpha \mkern-5mu = \mkern-5mu 1$.
	\section{Experiments}

\textbf{Data.}
We empirically evaluate our work on the real-world ETH Entomological Collection (ETHEC) dataset \cite{dhall_20.500.11850/365379} comprising images of \textit{Lepidoptera} specimens with their taxonomy tree. The real-world dataset has variations not only in terms of the images per category but also a significant imbalance in the structure of the taxonomical tree. In \cref{fig:ethec_distribution} we illustrate the data distribution for each label in the ETHEC hierarchy.

\begin{figure}[!htbp]
    \centering
    \includegraphics[width=0.375\textwidth]{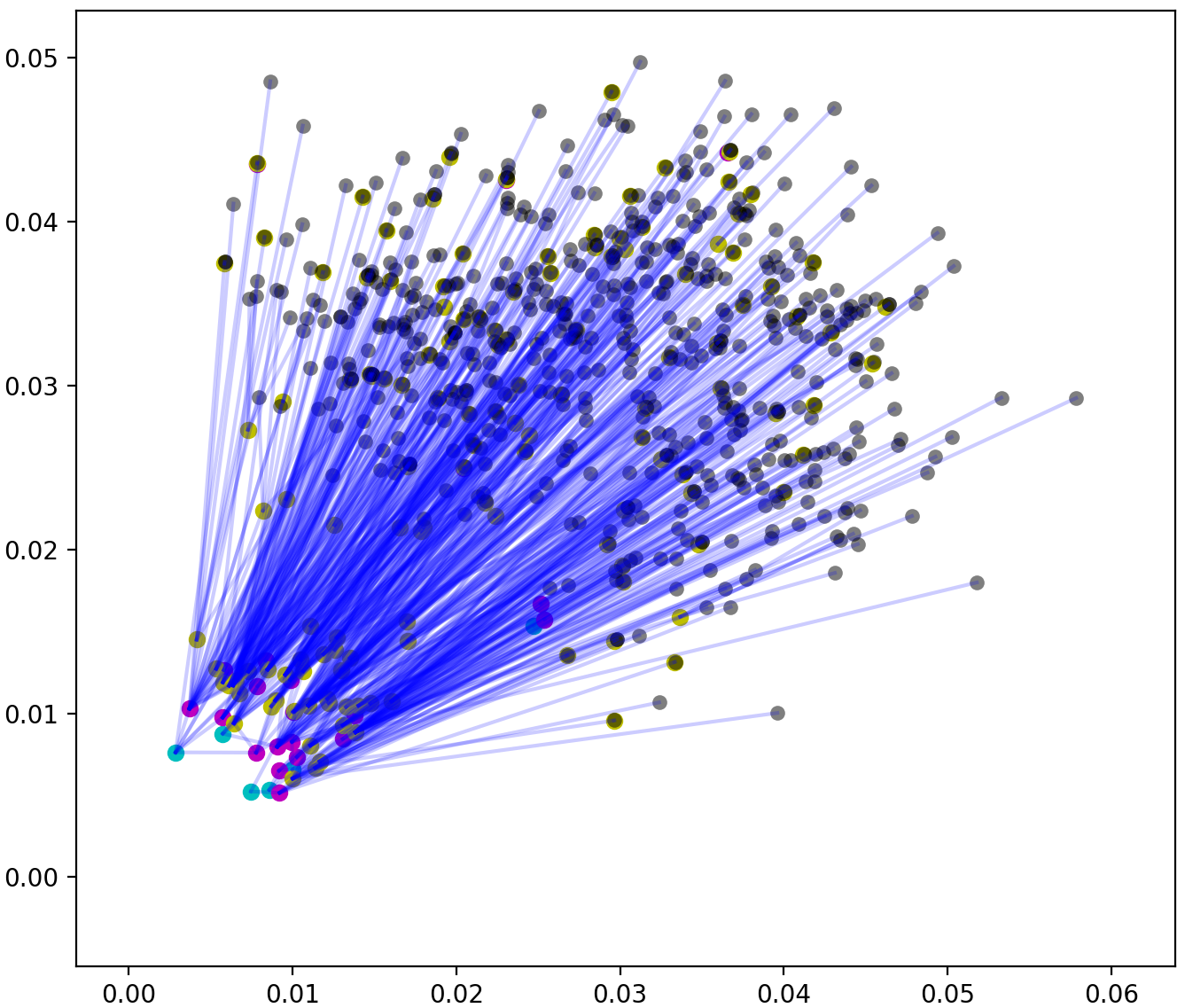}
    \vskip\baselineskip
    \caption{Label-only embeddings with HC $\mathbb{D}^{1000}$ projected to 2D. The embeddings organize themselves such that more generic concepts are closer to the origin while the most specific concepts form the periphery. Color coding as \cref{fig:oe_2d_labels}.} 
    \label{fig:hc_1000d}
    \vspace{-0.3cm}
\end{figure}

\begin{table}[!htbp]
\centering
\setlength\tabcolsep{3.5pt} 
\begin{tabular}{| c || c | c | c || c | c | c |}
  \hline
  & \multicolumn{3}{|c|}{classify \emph{test} set images} & \multicolumn{3}{|c|}{graph reconstruction} \\ \hline
  
  Model & \textbf{m-F1} & hit@3 & hit@5 & TPR & TNR & full-F1 \\ [0.5ex] 
  \hline\hline
  \multicolumn{7}{|c|}{Euclidean Cones} \\ \hline
  $d=10$ & 0.780 & 0.889 & 0.920 & 0.805 & 0.998 & 0.704 \\ \hline
  $d=10^2$ & 0.835 & 0.902 & 0.943 & \textbf{0.963} & \textbf{0.999} & \textbf{0.821} \\ \hline
  $d=10^3$ & 0.801 & 0.897 & 0.928 & 0.815 & 0.998 & 0.707 \\ \hline
  \hline
  \multicolumn{7}{|c|}{Hyperbolic Cones} \\ \hline
  $d=10^2$ & \textbf{0.840} & \textbf{0.920} & \textbf{0.939} & 0.642 & 0.998 & 0.576 \\ \hline
  $d=10^3$ & 0.805 & 0.902 & 0.928 & 0.523 & 0.997 & 0.483 \\
  \hline\hline

\end{tabular}
\caption{The table summarizes the embedding model performance when used to classify images for the ETHEC dataset \cite{dhall_20.500.11850/365379}. The joint image and label embeddings live in $\mathbb{R}^{d}$ or $\mathbb{D}^{d}$. m-F1 is the critical metric for image classification performance. We also report the quality of the reconstruction for the label-hierarchy after the joint embedding.}
\label{table:emb_classification_full_results}
\vspace{-0.5cm}
\end{table}

\subsection{Hierarchical Classification Performance}
To perform image classification using embeddings, the least violating energy $E(f_{l}(l), f_{i}(i))$ for a given image across all possible labels in a given level in the hierarchy is considered as the predicted label. The CNN models use Adam \cite{kingma2014adam} for 100 epochs with 224 x 224 RGB images and batch size=64. For HAB, PLC: $\text{lr} \mkern-5mu = \mkern-5mu 10^{-2}$; MC, M-PLC, HS: $\text{lr}  \mkern-5mu = \mkern-5mu 10^{-5}$. We empirically found ResNet-50 for HAB, PLC, MC, M-PLC and ResNet-152 for HS among ResNet 50, 101, 152 variants.

\begin{table}[!htbp]
\centering
\setlength\tabcolsep{4.0pt} 
\begin{tabular}{| c || c || c | c | c | c |} 
 \hline
   Model & m-F1 & $L_{1}$ & $L_{2}$ & $L_{3}$ & $L_{4}$  \\ [0.5ex] 
 \hline\hline
 \multicolumn{6}{|c|}{CNN-based methods} \\
 \hline \hline
  HAB & 0.8147 & 0.9417 & 0.9446 & 0.8311 & 0.4578 \\ \hline \hline
  PLC & 0.9084 & 0.9766 & 0.9661 & 0.9204 & 0.7704 \\ \hline
  MC & \underline{\textbf{0.9223}} & \underline{\textbf{0.9887}} & \underline{\textbf{0.9758}} & \underline{\textbf{0.9273}} & \underline{\textbf{0.7972}} \\ \hline
  M-PLC & 0.9173 & 0.9828 & 0.9701 & 0.9233 & 0.7930 \\ \hline
  HS & 0.9180 & 0.9879 & 0.9731 & 0.9253 & 0.7855 \\ \hline
  \multicolumn{6}{|c|}{Order-preserving (joint) embedding models} \\
  \hline \hline
  EC d=100 & 0.8350 & 0.9728 & 0.9370 & 0.8336 & 0.5967 \\ \hline
  HC d=100$*$ & 0.7627 & 0.9695 & 0.9205 & 0.7523 & 0.4246 \\ \hline
  HC d=100 & \textbf{0.8404} & \textbf{0.9800} & \textbf{0.9439} & \textbf{0.8477} & \textbf{0.5977} \\ \hline
  
\end{tabular}
\caption{Both EC and HC exploit hierarchical information and outperform the hierarchy-agnostic classifier baseline. We include the overall m-F1 in addition to the separate m-F1 across the 4 levels in the ETHEC dataset \cite{dhall_20.500.11850/365379}. All joint-embeddings models are initialized using labels-only embeddings. $*$=random initalization, \underline{best overall model}, \textbf{best model in category}.}
\label{table:combined_levelwise_performance}
\end{table}


\cref{table:combined_levelwise_performance} shows that the hierarchy-agnostic baseline is outperformed by all models that use any kind of hierarchical information. Embeddings: a completely different class of models, used widely in context of natural language but are relatively unexplored for image classification, also outperform HAB.

\begin{figure}[!htbp]
    \centering
    \includegraphics[width=0.47\textwidth]{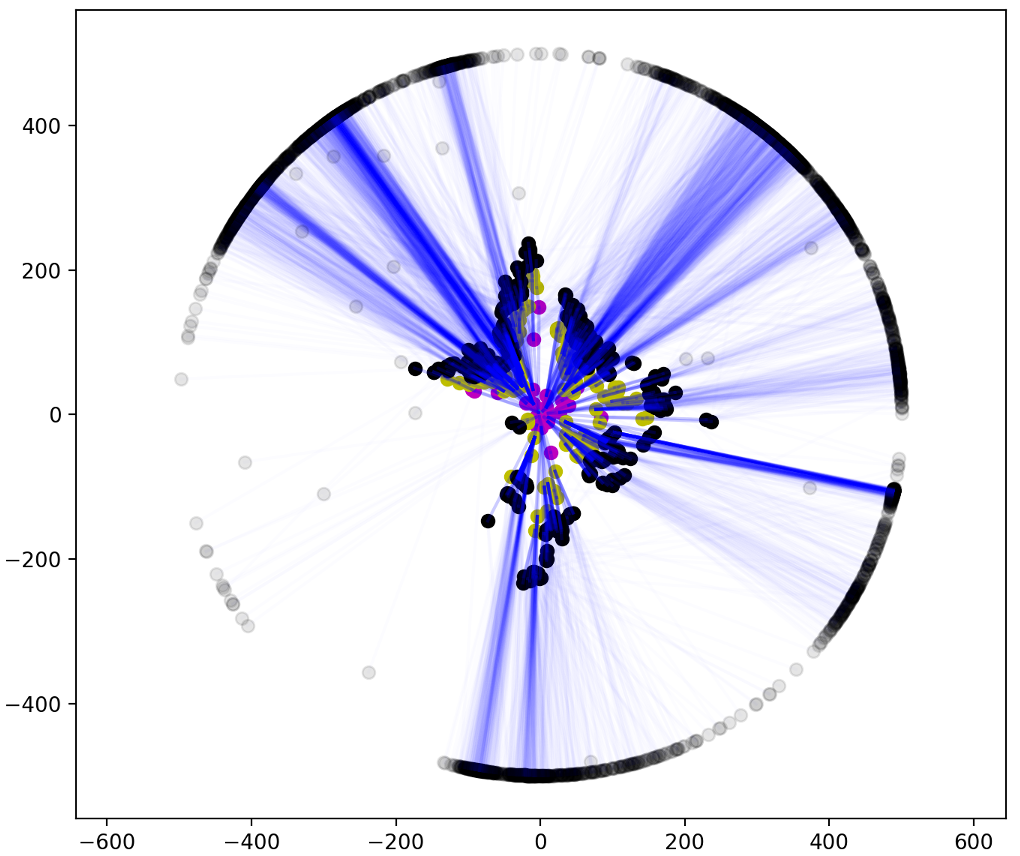}
    \caption{Jointly embedding labels and images using EC in $\mathbb{R}^2$. Color coding follow \cref{fig:oe_2d_labels}, grey: images. The images are accumulated around the periphery, away from the origin.}
    \label{fig:ec_2d_labels_images}
    \vspace{-0.5cm}
\end{figure}
\textbf{W's model capacity.} We use a matrix $W$ that transforms \textit{fc7} image features to the embedding space. A more elaborate 4-layer feed-forward neural network was also used but performed worse and was hard to optimize. Jointly training the complete CNN was also over-fitting.

\textbf{Negative edge frequency.} For joint-embedding the ETHEC dataset \cite{dhall_20.500.11850/365379}, since the images (around 50,000) outnumber the labels (723) we thought it might be useful to randomly sample negative edges such that the ratio of negative nodes have a proportion to be 50\%:50\% for images:label ratio however, the original strategy works better.

\textbf{Choice of Optimizer.} Initial experiments for the hyperbolic cones (HC) used the RSGD optimizer as it seemed to work for labels-only embeddings hyperbolic cones. When using the same to optimize over the labels for the joint-embedding model, we noticed that the label hierarchy moves towards the image labels and ends up collapsing from a very good initialization (taken from the labels-only embeddings). The collapse leads to entanglement between nodes from different labels and images, which leads it to a point of no return and the performance worsens due to the label-hierarchy becoming disarranged and its inability to recover. We believe that the reason for its inability to rearrange is due to there being a two different types of objects being embedded (and also being computed differently) and it compounded by using different optimizers.

In our experiments we obtain best results when using the Adam optimizer even if it means the update step for parameters living in hyperbolic space has to be performed in an approximate manner. Adam optimizer with an approximate update step works better in practice than RSGD with its mathematically more precise update step.

\textbf{Label initialization for joint-embeddings}
Using RSGD we observed that if the labels are not initialized with the labels-only embedding then the joint model finds it difficult to disentangle the label embeddings and eventually this effect is cascaded to the images causing the image classification performance to not improve.

With the RSGD replaced by the Adam optimizer, in experiments where we randomly initialized the label-embeddings, we observed them to disentangle and form entailment cones even with the images being involved and making the optimization more complex. The joint-model still works well with random label initialization and achieves an image classification m-F1 score of 0.7611 and even outperforms the hierarchy-agnostic CNN in the m-F1 $L_1$. \cite{ganea2018entailment_cones} recommends to use Poincar\'e embeddings \cite{nickel2017poincare} to initialize the hyperbolic cones model. The fact that the joint model as well as the labels-only hyperbolic cones have great performance without any special initialization scheme is interesting. We conjecture that this could be because of using an approximate yet better optimizer. 

	\section{Conclusion}
We propose an embedding-based approach for image classification using \emph{entailment cones}, a recently proposed type of \emph{order-preserving} embeddings. In particular, we compare these both in the Euclidean geometry setting and in the hyperbolic setting, and show that hyperbolic geometry provides an empirical advantage over Euclidean geometry. We also propose and compare a set of simple hierarchical classifier baselines where the hierarchy is incorporated in the loss function. Although these tend to perform slightly better than embedding-based approaches, they are less flexible as they assume that the hierarchy is fixed, and are more limited in terms of downstream tasks (e.g.\ they do not allow for hierarchy-based retrieval). Finally, we evaluate our methods on the real-world ETHEC dataset \cite{dhall_20.500.11850/365379}, and show that exploiting hierarchical information always leads to an improvement over a shallow CNN classifier.

    
    \bibliographystyle{ieeetr}
	\bibliography{refs}

    \clearpage
\section{Appendix}
\subsection{Schematics for CNN-based models}


\subsection{Performance metrics}
\paragraph{True positive rate}
True positive rate (TPR) is the fraction of actual positives predicted correctly by the method.
\begin{equation}
    \text{TPR}=\frac{tp}{totalPositives}
\label{eq:TPR-formula}
\end{equation}
\paragraph{True negative rate}
True negative rate (TNR) is the fraction of actual negatives predicted correctly by the method.
\begin{equation}
    \text{TNR}=\frac{tn}{totalNegatives}
\label{eq:TNR-formula}
\end{equation}
\paragraph{Precision}
Precision computes what fraction of the labels predicted true by the model are actually true.
\begin{equation}
    \text{P}=\frac{tp}{tp+fp}
\label{eq:precision-formula}
\end{equation}
\paragraph{Recall}
Recall computes what fraction of the true labels were predicted as true.
\begin{equation}
    \text{R}=\frac{tp}{tp+fn}
\label{eq:recall-formula}
\end{equation}
\paragraph{F1-score}
\begin{equation}
    \text{F1}=\frac{2*P*R}{P+R}
\label{eq:f1-formula}
\end{equation}
\paragraph{Hit@k}
\begin{equation}
    \text{Hit@K}=\frac{1}{N} \sum_{i=1}^{N} 1[\text{label}_{i}^{\text{gt}} \in \text{SortedPredictions}(i)]
\label{eq:hit-at-k-formula}
\end{equation}
where, $\text{SortedPredictions}(i) = \{ \text{label}_{0}^{\text{pred}}, \text{label}_{1}^{\text{pred}},..., \text{label}_{k-1}^{\text{pred}}, \text{label}_{k}^{\text{pred}}\}$ is the set of the top-K predictions for the \textit{i}-th data sample.
\paragraph{Macro-averaged score}
A macro-averaged score for a metric is calculated by averaging the metric across all labels.
\begin{equation}
    \text{M-metric}=\frac{1}{N} \sum_{i=1}^{N} \text{metric}(\text{label}_{i})
\label{eq:macro-metric-formula}
\end{equation}

\paragraph{Micro-averaged score}
A micro-averaged score for a metric is calculated by accumulating contributions (to the performance metric) across all labels and these accumulated contributions are used to calculate the micro score.

\subsection{ETHEC dataset}
The ETHEC dataset \cite{dhall_20.500.11850/365379} contains 47,978 images of the ``order'' \textit{Lepidoptera} with corresponding labels across 4 different levels. According to the way the taxonomy is defined, the \textit{specific epithet} (species) name associated with a specimen may not be unique. For instance, two samples with the following set of labels, (\textit{Pieridae}, \textit{Coliadinae}, \textit{Colias}, \textit{staudingeri}) and (\textit{Lycaenidae}, \textit{Polyommatinae}, \textit{Cupido}, \textit{staudingeri}) have the same \textit{specific epithet} but differ in all the other label levels - \textit{family}, \textit{subfamily} and \textit{genus}. However, the combination of the \textit{genus} and \textit{specific epithet} is unique. To ensure that the hierarchy is a tree structure and each node has a unique parent, we define a version of the database where there is a 4-level hierarchy - \textit{family} (6), \textit{subfamily} (21), \textit{genus} (135) and \textit{genus + specific epithet} (561) with a total of 723 labels. We keep the \textit{genus} level as according to experts in the field, information about genera helps distinguish among samples and result in a better performing model.

\subsection{HAB details}
\label{subsec:hab_details}
Here we discuss the details of having a single threshold for every label or a common threshold for all labels in a multi-label classification setting. Here we observe the maximum and minimum labels predicted by the multi-label model across the whole dataset. We also look at the mean and standard deviation of the number of labels predicted.
\subsubsection{Per-class decision boundary (PCDB) models}
The ill-effects of such free rein are reflected in \cref{table:ethec_merged_ft_multi_label}. Models with a high average number of predictions, especially the per-class decision boundary (PCDB) models, have high recall as they predict a lot more than just 4 labels for a given image. Predicting the image's membership in a lot of classes improves the chances of predicting the correct label but at the cost of a large number of false positives. The (min, max), $\mu \pm \sigma$ column clearly shows the reckless behavior of the model predicting a maximum of 718 labels for one such sample and 451.14 $\pm$ 136.69 on average for the worst performing multi-label model in our experiments.

\subsubsection{One-fits-all decision boundary (OFADB) models}
The one-fits-all decision boundary (OFADB) performs better than the same model with per-class decision boundaries (PCDB). We believe that the OFADB prevents over-fitting, especially in the case when many labels have very few data samples to learn from, which is the case for the ETHEC database. Here too, the nature of the multi-label setting allows the model to predict as many labels as it wants however, there is a marked difference between the (min, max), $\mu \pm \sigma$ statistics when comparing between the OFADB and PCDB. The best performing OFADB model predicts 3.10 $\pm$ 1.16 labels on average. This is close to the correct number of labels per specimen which is equal to the 4 levels in the label hierarchy.

\subsubsection{Loss reweighing and Data re-sampling}
Both data re-sampling and loss re-weighing remedy imbalance across different labels but via different paradigms. Instead of modifying what the model sees during training, reweighing the loss instead penalizes different data points differently. We choose to use the inverse-frequency of the label as weights that scale loss corresponding to the data point belonging to a particular label.

re-sampling involves choosing some samples multiple times while omitting others by over-sampling and under-sampling. We wish to prevent the model from being biased by the population of data belonging to a particular label. We perform re-sampling based on the inverse-frequency of a label in the \emph{train} set. In our experiments re-sampling significantly outperforms loss reweighing confirming the observations made in \cite{seiffert2008resampling}.

\begin{table}[!htbp]
\centering
\setlength\tabcolsep{3.0pt} 
\begin{tabular}{| c | c || c | c | c || c |} 
 \hline
  cw & rs & m-P & m-R & m-F1 & (min, max), $\mu \pm \sigma$ \\ [0.5ex] 
 \hline\hline
 \multicolumn{6}{|c|}{ResNet-50 - Per-class decision boundary} \\
 \hline\hline
  
  \ding{55} & \ding{55} & 0.0355 & 0.7232 & 0.0677 & (3, 351), 81.4 $\pm$ 69.5 \\
  \ding{55} & \ding{51} & 0.7159 & 0.7543 & \textbf{0.3718} & (0, 13), 4.2 $\pm$ 2.1 \\
  \ding{51} & \ding{55} & 0.0077 & \textbf{0.8702} & 0.0153 & (84, 718), 451.1 $\pm$ 136.7 \\
  \ding{51} & \ding{51} & 0.0081 & 0.7519 & 0.0161 & (33, 714), 370.0 $\pm$ 120.6 \\
 \hline\hline
 \multicolumn{6}{|c|}{ResNet-50 - One-fits-all decision boundary} \\
 \hline\hline
  \ding{55} & \ding{55} & 0.9324 & 0.7235 & \textbf{0.8147} & (0, 7), 3.1 $\pm$ 1.2 \\
  \ding{55} & \ding{51} & \textbf{0.9500} & 0.6564 & 0.7763 & (0, 5), 2.8 $\pm$ 0.6 \\
   \ding{51} & \ding{55} & 0.2488 & 0.2960 & 0.2704 & (4, 9), 4.8 $\pm$ 0.8 \\
  \ding{51} & \ding{51} & 0.1966 & 0.3800 & 0.2591 & (4, 10), 7.7 $\pm$ 0.6 \\
 [1ex] \hline
\end{tabular}
\caption{Performance metrics for the HAB on the ETHEC dataset. The models used in this experiment are pre-trained on the 1000-class ImageNet data set. All weights are updated with a learning rate of 0.01, a batch-size of 64 and input spatial dimensions are 224x224 for 100 epochs. \textit{P}, \textit{R} and \textit{F1} represent Precision, Recall and F1-score; \textit{cw} and \textit{rs} represent class weight and re-sampling. \textit{m} are micro-averaged metrics. The top performing models are in bold-face. Since, the model can predict any number of labels (between 0 and $N_{total}$), the table includes the minimum and the maximum number of labels predicted \textit{(min, max)} as well as the number of labels predicted on average $\mu \pm \sigma$. These statistics, like the rest, are calculated for samples in the \emph{test} set.}
\label{table:ethec_merged_ft_multi_label}
\end{table}

\begin{figure*}[!htbp]
    \centering
    \begin{subfigure}[b]{0.480\textwidth}
        \centering
        \includegraphics[width=\textwidth]{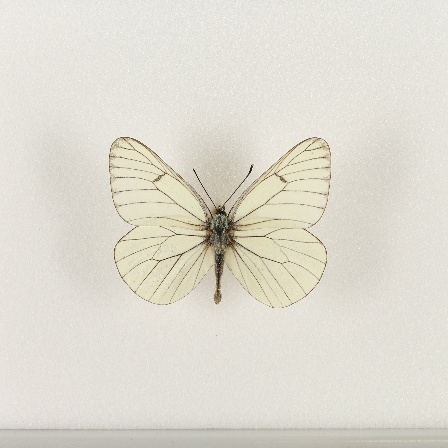}
        \caption[]%
        {\textit{Aporia crataegi} [ENT01\_2017\_03\_27\_007897]}    
        \label{fig:ac1}
    \end{subfigure}
    \hfill
    \begin{subfigure}[b]{0.480\textwidth}  
        \centering 
        \includegraphics[width=\textwidth]{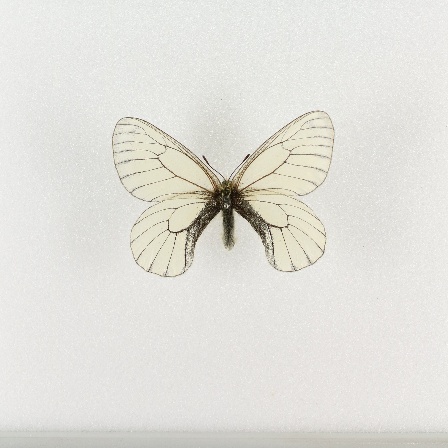}
        \caption[]%
        {\textit{Parnassius stubbendorfii} [ENT01\_2018\_03\_09\_132877]}    
        \label{fig:ps1}
    \end{subfigure}
    \vskip\baselineskip
    \begin{subfigure}[b]{0.480\textwidth}   
        \centering 
        \includegraphics[width=\textwidth]{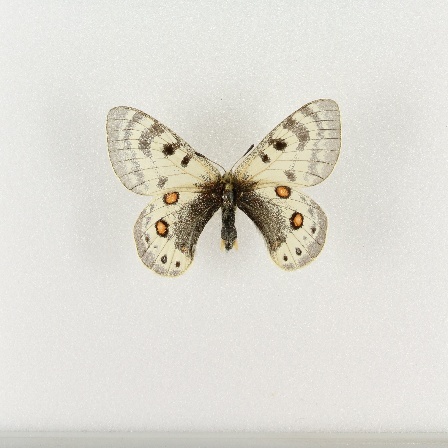}
        \caption[]%
            {\textit{Parnassius delphius} [ENT01\_2018\_03\_09\_133076]}    
        \label{fig:pd1}
    \end{subfigure}
    \quad
    \begin{subfigure}[b]{0.480\textwidth}   
        \centering 
        \includegraphics[width=\textwidth]{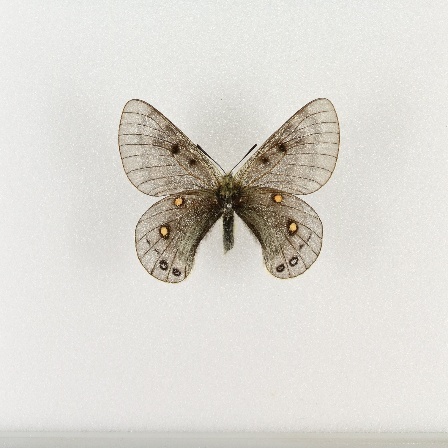}
        \caption[]%
            {\textit{Parnassius delphius} [ENT01\_2018\_03\_09\_133091]}
        \label{fig:pd2}
    \end{subfigure}
    \caption[]
    {Both semantic similarity and visual similarity are required to perform tasks relating to image understanding. Here, we see an example from the ETHEC dataset \cite{dhall_20.500.11850/365379}. At first glance, (a) and (b) look like they belong to the same class and so do (c) and (d) considering the visual similarities. However, this is not so straight-forward as (a) and (b) belong to two separate genera and species but have a really low inter-class variance. On the other hand, (b), (c) and (d) all share the same genus \textit{Parnassius} but have a larger intra-class variance than (a) and (b). This demonstrates how visual similarity might not imply semantic similarity and vice-versa.} 
    \label{fig:semantics-vs-visuals}
\end{figure*}

\begin{figure*}[!htbp]
    \centering
    \begin{subfigure}[!h]{0.49\textwidth}
        \centering
        \includegraphics[width=\textwidth]{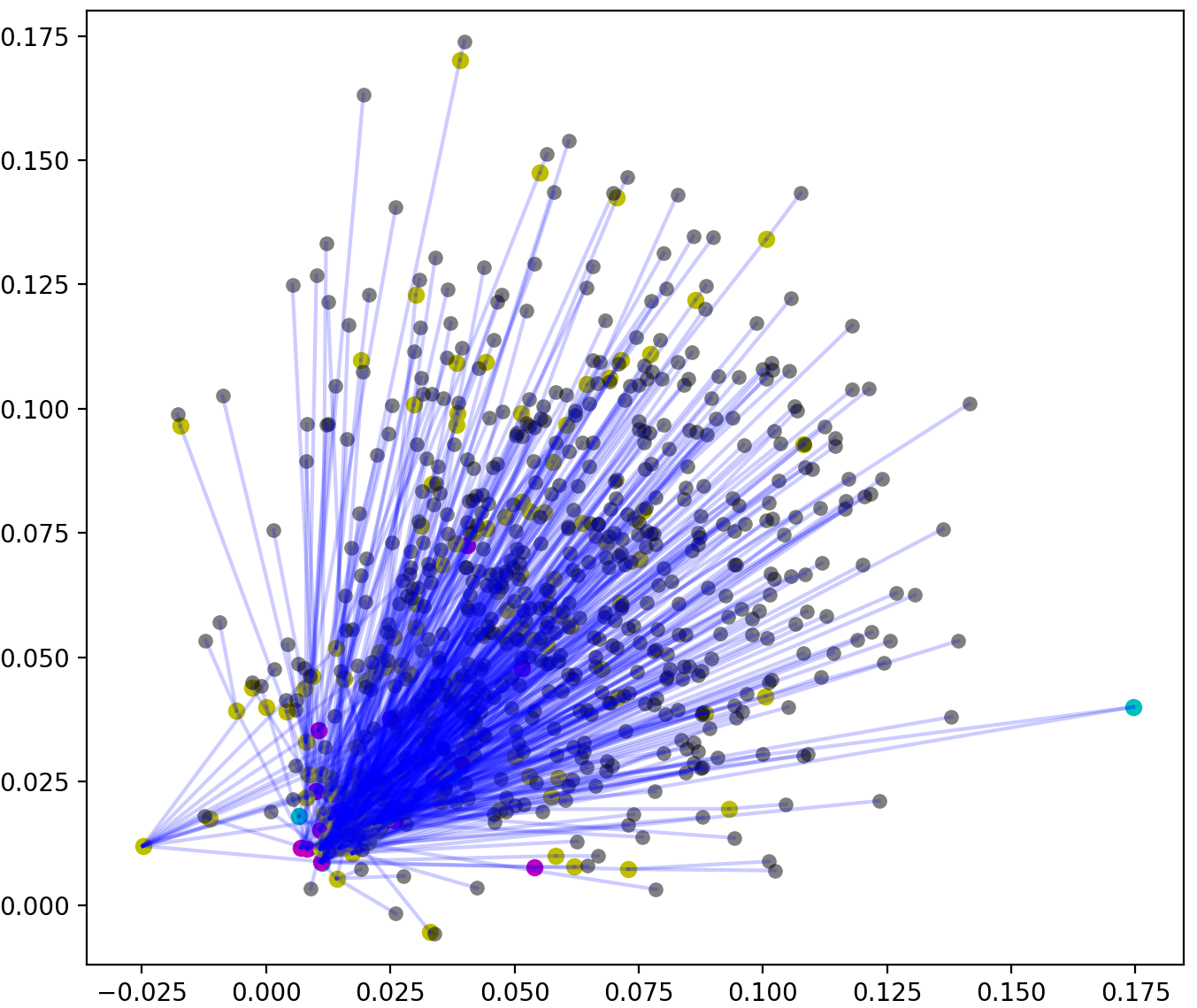}
        \caption[]%
        {{\small Hyperbolic Cones 100-D}}    
        \label{fig:hc_100d_labels}
    \end{subfigure}
    \hfill
    \begin{subfigure}[!h]{0.49\textwidth}  
        \centering 
        \includegraphics[width=\textwidth]{images/appendix/hc/1000d_4999.png}
        \caption[]%
        {{\small Hyperbolic Cones 1000-D}}    
        \label{fig:hc_1000d_labels}
    \end{subfigure}
    \vskip\baselineskip
    \caption[]
    {Projected visualization of labels embedded using hyperbolic cones in 100 and 1000 dimensions. The cyan nodes represent \emph{family}, the magenta nodes represent \emph{sub-family}, the yellow nodes \emph{genus} and black nodes \emph{genus+species}. This resembles a flower-like shape where the more generic concepts are closer to the origin and at the base of this flower-like shape and most specific concepts at the tip of the petals which forms the periphery are a visible the most (=black nodes).} 
    \label{fig:hc_100d_1000d}
\end{figure*}

  \end{document}